\documentclass[a4paper, 10pt, conference]{ieeeconf}      

\IEEEoverridecommandlockouts                              

\usepackage{mathrsfs}
\usepackage{graphicx}
\usepackage{cite}
\usepackage{bm}
\usepackage{color}
\usepackage{amsmath,amssymb,amsfonts}
\usepackage{algorithmic}
\usepackage{algorithm2e}
\usepackage{graphicx}
\usepackage{textcomp}
\usepackage{xcolor}
\usepackage{comment}  
\usepackage{url} 
\usepackage[utf8]{inputenc} 
\usepackage[T1]{fontenc}    
\usepackage{booktabs}
\def\BibTeX{{\rm B\kern-.05em{\sc i\kern-.025em b}\kern-.08em
    T\kern-.1667em\lower.7ex\hbox{E}\kern-.125emX}}

\usepackage[hang,small,bf]{caption}
\usepackage[subrefformat=parens]{subcaption}
\newcommand{\argmin}{\mathop{\rm arg~min}\limits}

\newcommand{\mathS}{\mathcal{S}}

\newcommand{\mathE}{\mathcal{E}}
\newcommand{\mathF}{\mathcal{F}}

\newcommand{\rad}{\mathrm{rad}}

\usepackage{amsthm}
\theoremstyle{definition}
\newtheorem{definition}{Definition}

\usepackage[colorlinks,bookmarks=false]{hyperref}

\title{\LARGE \bf
Discovering Avoidable Planner Failures of Autonomous Vehicles\\ using Counterfactual Analysis in Behaviorally Diverse Simulation
}

\author{Daisuke Nishiyama$^{1}$, Mario Ynocente Castro$^{1}$, Shirou Maruyama$^{1}$, Shinya Shiroshita$^{1}$, Karim Hamzaoui$^{1}$,\\
Yi Ouyang$^{2}$, Guy Rosman$^{3}$, Jonathan DeCastro$^{3}$, Kuan-Hui Lee$^{3}$ and Adrien Gaidon$^{3}$
\thanks{$^{1}$Preferred Networks, Inc., Japan.,
        {\{dnishiyama, marioyc, maruyama, shiroshita,  karim\}@preferred.jp}}%
\thanks{$^{2}$Preferred Networks America, Inc., U.S.,
        {ouyangyi@preferred-america.com}}%
\thanks{$^{3}$Toyota Research Institute, U.S.,
        {\{firstname.lastname\}@tri.global}}%
}

\begin{document}

\maketitle
\thispagestyle{empty}
\pagestyle{empty}

\begin{abstract}
Automated Vehicles require exhaustive testing in simulation to detect as many safety-critical failures as possible before deployment on public roads. In this work, we focus on the core decision-making component of autonomous robots: their planning algorithm. We introduce a planner testing framework that leverages recent progress in simulating behaviorally diverse traffic participants. Using large scale search, we generate, detect, and characterize dynamic scenarios leading to collisions. In particular, we propose methods to distinguish between unavoidable and avoidable accidents, focusing especially on automatically finding planner-specific defects that must be corrected before deployment. Through experiments in complex multi-agent intersection scenarios, we show that our method can indeed find a wide range of critical planner failures.
\end{abstract}

\section{Introduction}

Thorough verification of planning algorithms is critical to show their robustness in handling various real-world situations, especially for automated vehicles interacting with other traffic participants.

As the planner is responsible for making decisions autonomously, its defects are likely to cause accidents.
Therefore, exhaustive and reproducible testing in safety-critical scenarios using simulated environments is a core part of the development of automated vehicles.
In recent years, various traffic simulators for autonomous driving were proposed~\cite{carla, sumo, fluids}. 
Benchmarks such as CommonRoad~\cite{common_road} also provide various log data of vehicle behaviors for use in evaluation. However, these environments typically lack diversity in the behavior of traffic participants, thus limiting test coverage.
On the other hand, increasing diversity by introducing random driving policies or adversarial vehicles results mostly in unpredictable cases~\cite{adversarial-test, adaptive-test1}. This limits testing to situations where the optimal planning behavior is overly conservative or mandates drastic evasive maneuvers, leaving most realistic safety-critical situations untested.

In this work, we explore \emph{how AI agents with diverse RL policies can help find a wide range of interesting planner failure modes}.

Our \textbf{first contribution} is a large scale dynamic scenario generation strategy leveraging recent progress in Reinforcement Learning (RL) for driving agents~\cite{diversity-rich-sim}. Using behaviorally diverse policies along with a flexible action space, we efficiently and automatically generate test cases covering a wider range of driving behaviors than what typical rule-based approaches can.
The resulting scenarios enable discovering many situations resulting in safety-critical failures, especially collisions with other simulated traffic participants. However, not all collisions might be meaningful or useful for planner testing.
Therefore, in our \textbf{second and main contribution}, we focus on counterfactually detecting avoidable failures that can (and should) be fixed. Unavoidable failures are situations where a planner cannot escape a collision no matter how it behaves, and thus are an intrinsic feature of the environment, not the planning algorithm itself.
Following a collision, we first rewind the traffic simulation back in time, considering human reaction time as a benchmark~\cite{driver-reaction-time}.
We then propose two methods for assessing, counterfactually, if a failure is avoidable or not: a \emph{planner-specific} one and a \emph{generic} one.
Our planner-specific failure detection method uses general parameter optimization tools (e.g. \cite{optuna}) to try to find a set of planner parameters that could have enabled it to avoid the observed failure, which doubles as useful feedback for planner improvements.
Our generic avoidability check estimates whether the collision is dynamically avoidable by replacing the target planner with a collision-avoiding agent trained using a Constrained Markov Decision Process (CMDP) approach~\cite{maeda2019reconnaissance}. This method can be applied when searching for planner parameters is not possible.
In both cases, we check if the failure could have been avoided by resuming the simulation. This is different from historical replay, as all agent policies are rerun.

To experimentally validate our method, we test the standard "Intelligent Driver Model" (IDM)~\cite{idm}, modified to drive in multiple intersection scenes where the other agents use the aforementioned diverse RL policies.
Our experiments reveal significant defects covering a wide range of near-misses and collision types, angles, and positions, thus confirming the usefulness of our approach for improving the safety of planning algorithms.

\section{Related Work}
With the recent progress in the development of autonomous vehicles, studies focusing on testing planner modules have become an increasing focus of research in recent years.
Due to the difficulty of actual vehicle testing and evaluation of autonomous driving systems~\cite{karla2016}, various approaches using simulation testing have been proposed.

Tuncali {\it et al.}~\cite{adversarial-test} proposed an adversarial error detection method for ML-based autonomous vehicles modules, such as planner and recognition modules, by perturbing test scenarios to induce collisions and misclassifications.
O'Kelly {\it et al.}~\cite{rare-event-sim} proposed an importance sampling and  a cross-entropy method for a given probability distribution for generating test scenarios in which more accidents occur.
In \cite{rare-event-sim, zhao2017, huang2017}, naturalistic driving data were used for making the behavior model of other vehicles. However, collecting data might not always be possible in rare or safety-critical scenarios.

Koren {\it et al.}~\cite {adaptive-test1} tackled the problem of finding collision failures for the IDM-based planner, using a solver to sample pedestrian behavior across crosswalks. They showed experimentally that collisions could be induced more efficiently using a deep RL-based method, and compared it to using Monte Carlo Tree Search (MCTS).
Corso {\it et al.}~\cite {adaptive-test2} further improved Koren {\it et al.}'s work by proposing a method for detecting failure cases that are relevant to the planner. Their method evaluates the planner's behavior with a Responsibility-Sensitive Safety (RSS) condition~\cite{rss}, which focuses on the faults caused by the planner.
They separately tried to diversify the failure cases the method detects by giving a reward to the solver when the planner's trajectory on a new failure is far from the previously found collision trajectories.

Corso {\it et al.}'s motivation is similar to our purpose. However, they only considered a planner with other pedestrians instead of vehicles, and they used the hard-coded RSS condition while our work introduces a more flexible and relevant concept of avoidable failures. Furthermore, they have only considered either finding diverse failure trajectories or failures relevant to the tested planner, while we focus on both aspects at the same time.
\section{Generating diverse planner test cases}

\subsection{Traffic Simulator Design}

\begin{figure}[t!]
    \centering
    \includegraphics[width=7cm]{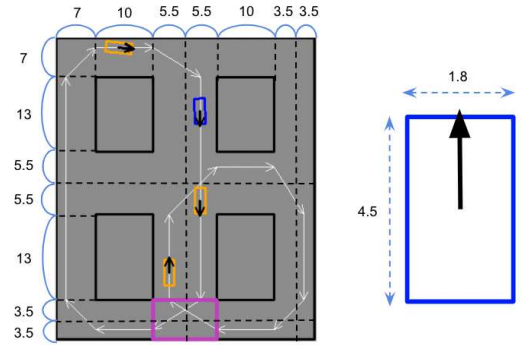}
    \caption{
    The scale of the traffic simulation map and the vehicle.
    All units are meters.
    The black lines in the map represent walls.
    The black arrows of vehicles mean headings.
    }
    \label{fig:map-and-vehicle}
\end{figure}

We use a testing environment based on a traffic simulator for intersection scenes used in the work of \cite{diversity-rich-sim}.
Fig.~\ref{fig:map-and-vehicle} shows a map of the intersection scene and the scales used for roads and vehicles.
Vehicles are limited from going over the black lines, representing walls, and overlapping on each other.

The traffic simulator has one ego vehicle (blue) $g^e$ as the planner to be evaluated, 
and $m$ other vehicles (orange) $g^{o_1}, \dots, g^{o_m}$ which can be set.
The ego vehicle computes the next action from the current state according to a planner policy $\pi^p_\theta$, where $p$ indicates the \emph{planner}'s policy to distinguish it from the other vehicle policies described below,
$\theta$ is a combination of adjustable parameters of the planner,
and users can select any parameters from a defined parameter space $\Theta_p$.
Also, each other vehicle computes the next action according to a policy $\pi^o$. The user can select any policy from a set of available other vehicle policies $\Pi_o$.
The simulator updates the state at discrete time intervals with a constant defined by the user.
A single update to the states in the simulator is called a \emph{step},
and $\Delta t$ is the update interval time constant for one step.
The action of the vehicle $g^k$ at step $i$ is represented by a pair $a_i^k = (\alpha, \phi)$ of acceleration $[m/s^2]$ and steering value $[\rad]$.
The state of the vehicle $g^k$ at step $i$ is represented as a quadruplet $u^k_i = (x^k_i, y^k_i, v^k_i, h^k_i)$, including $(x^k_i, y^k_i)$, the position of the center in x-y coordinates~$[m]$, $v^k_i$, the velocity~$[m/s]$, and $h^k_i$, the heading~$[\rad]$.

Each vehicle $g^k$ transitions from the current state $u_i^k$ to the next state $u_{i+1}^k$ from the selected action of the policy according to the kinematic bicycle model~\cite{bicycle-model}.
The state of the simulator in step $i$ is defined as $U_i = (u_i^e, u_i^{o_1}, \dots, u_i^{o_m})$, where $u_i^e$ and $u_i^{o_k}$ are a state of the ego vehicle $g^e$ and one of the other vehicles $g^{o_k}$ at step $i$, respectively.
The users can define a set of failure states $\mathF$ for all failure conditions of the planner vehicle.
The user can define a set of end states $\mathE$ to determine the termination of the simulation.
Unless otherwise noted, $\mathF \subseteq \mathE$ so the simulation will be terminated when it enters a failure state.

The simulator is given as inputs a planner policy $\pi^p_{\theta}$ and a \emph{scenario} $s = (\pi^o, U_0, \mathF, \mathE)$ which is an assortment of information on everything other than $\pi^p_\theta$, and then starts the simulation.
The set of all evaluation scenarios prepared by the user is represented as $\mathS$.
We also assume that all action choices and state transitions are deterministic for reproducibility.
That is, given ($\pi^p_{\theta}, s)$ to the simulator,
the state $U_n$ after $n$ steps is uniquely determined from $(\pi^p_\theta, s)$.
We denote as $T(\pi^p_{\theta}, s, n)$ the state derived by a ($\pi^p_{\theta}, s)$ after $n$ steps from the initial state of the simulation, and as $T(\pi^p_{\theta}, s, *)$ the last state of the simulation derived from $(\pi^p_{\theta}, s)$.
An input ($\pi^p_{\theta}, s$) is called a \emph{failure case} if the simulation ends in a state $T(\pi^p_{\theta}, s, *) \in \mathF$.
On the other hand, ($\pi^p_{\theta}, s$) is called a $\emph{success case}$ when the simulation ends in a state $T(\pi^p_{\theta}, s, *) \in \mathE \cap \mathF^{c}$.

\subsection{Generating Other Vehicle Policy Sets}\label{subsec:generating-other-vehicle-policy}

Using the method described in  \cite{diversity-rich-sim},
we generated RL-based driving policies for other vehicles.
Each vehicle is given a navigation route in advance, which defines a reward function for driving with the following goals: 
i) moving along the given navigation route as much as possible, 
and ii) not colliding to walls and vehicles.
During training, vehicles in the simulation $g^1, \dots, g^m$ move according to a policy $\pi^o$.
Note, however, that $\pi^o$ does not control all the vehicles in a centralized manner, but rather each of vehicles has a copy of $\pi^o$ which they execute in a decentralized manner.
Therefore, at the planner testing phase, it is possible to replace one of $g^1, \dots, g^m$ with a vehicle $g^e$ that drives according to a planner policy $\pi^p_\theta$ and execute the testing simulation.
The RL-based policies not only control the speed on the predefined path but also enable lateral control by steering.
This degree of freedom plays an important role in detecting diverse failures.

To acquire a diverse policy set, we use Diverse Policy Selection (DPS)~\cite{diversity-rich-sim} to generate $k$ policies with a success rate of $\delta$\% or more on the evaluation scenarios. See the original paper for details.

\subsection{Policy Set Diversity Metric}\label{subsec:diversity-metric}
We use a diversity metric inter-policy diversity~\cite{diversity-rich-sim} to evaluate how different the behaviors of any two policies in a set $\Pi_o$ are from each other.
Note that this $\Pi_o$ is used as a set of other vehicle policies during planner testing, but each policy $\pi^o$ in $\Pi_o$ is also treated as $\pi^o = \pi^p$ when evaluating this metric.
This evaluation for the policy set $\Pi_o$ is performed on a set of evaluation scenarios $\mathS$ which is different from training scenarios.

Let $\tau_s(\pi)$ be a trajectory generated from $(\pi, s)$,
and $\mathS_\pi$ a subset of $\mathS$ consisting of scenarios such that $(\pi, s)$ is a success case after running the simulation for a $\pi$ and all $s \in \mathS$.
The inter-policy diversity is defined as follows:
\begin{equation}
    D_{IP}(\Pi_o) = \frac{1}{|\Pi_o| (|\Pi_o| - 1)} \sum_{\pi\in \Pi_o}\sum_{\pi'\in \Pi_o\setminus\{\pi\}} D_{IP}(\pi, \pi')
\end{equation}
where
$D_{IP}(\pi, \pi') = \frac{1}{|\mathS_\pi \bigcap S_{\pi'}|} \sum_{s \in \mathS_\pi \bigcap S_{\pi'}} d(\tau_{s}(\pi), \tau_s(\pi')).$
Here $d(\tau,\tau')$ refers to the average Euclidean distance between two trajectories.
In planner testing, we use this diversity metric to quantify the diversity of different sets of policies generated for other vehicles.

\section{Avoidable Failures}

This section defines failure cases that are relevant to our study.
Intuitively, we only focus on failure cases which can be avoided by modifying the planning policy. When the planner cannot be modified in a way that allows it to avoid the failure, such a failure case is likely irrelevant to the planner design because no improvement can be made in this situation. On the other hand, avoidable failures could serve as valuable references when designing pre-crash safety systems.

Consider a failure case with a planner policy $\pi^p_\theta$ and a scenario $s = (\pi^o, U_0, \mathF, \mathE)$ such that $T(\pi^p_\theta, s, n) \in \mathF$ for some step $n$ before the simulation ends.
The idea is to conduct a counterfactual analysis by re-running the simulation with a modified planner policy $\pi^{p\prime}$ to see if it is possible to avoid the failure.
However, we do not want to restart the simulation too far back in time from the $n$-th step when the failure occurs because it could completely change the simulation trajectory of the target planner.
Therefore, we re-run the simulation only shortly before the failure state to evaluate this pre-failure situation. More specifically, let $\rho$ be the minimum reaction time of the ego vehicle, the simulation is then restarted from the step $n-\lceil\rho/\Delta t\rceil$.

As choices for the modified policy $\pi^{p\prime}$, we propose two types of counterfactual modification: a planner-specific modification and a generic modification independent of the planner. These two types of modification provide two classes of avoidable failures for planner testing, discussed in the following two subsections.

\subsection{Planner-specific Avoidable Failure}
\label{subsec:planner-dependent-AF}
A planner-specific modification is where the parameter $\theta$ of the original policy is modified to a different parameter $\theta^{\prime}$. The modified policy becomes $\pi^{p\prime} = \pi^p_{\theta'}$, and
a planner-specific avoidable failure is then defined as follows.

\begin{definition}[planner-specific avoidable failure]
Given a planner policy $\pi_{\theta}^p$ and a scenario $s = (\pi^o, U_{0}, \mathF, \mathE)$ such that $T(\pi_{\theta}^p, s, n) \in \mathF$,
$(\pi_{\theta}^p, s)$ is a \emph{planner-specific avoidable failure} if and only if there exists a parameters $\theta^{\prime} \in \Theta_p\setminus\{\theta\}$ such that $T(\pi^p_{\theta^{\prime}}, s^{\prime}, *) \in \mathE \cap \mathF^{c}$ where $s^{\prime} = (\pi^o, T(\pi_{\theta}^p, s, n - \lceil\rho/\Delta t\rceil), \mathF, \mathE)$.
\end{definition}

The avoidable failures defined here are dependent on the planner's specifications.
Such a planner-specific avoidable failure is highly relevant to the planner because it highlights the incapability of the planner to choose the appropriate parameter $\theta^{\prime}$ to avoid the upcoming failure.
On the other hand, other failure cases are of lesser usefulness since the planner cannot avoid them only by adjusting its parameter.
This modified parameter $\theta^{\prime}$ found in the test can be provided as useful feedback to planner development. 

\subsection{Generic Avoidable Failure}
\label{subsec:planner-independent-AF}

Instead of searching within the planner parameter space for a combination that could possibly avoid the failure in question, another approach is to verify if the failure is avoidable following some "safe" driving policy. If a failure is avoidable by the "safe" policy, the failure case and the way it can be avoided provide valuable information for improving the planner design. 

One intuitive candidate for the "safe" policy is to opt for a full deceleration in order to stop the vehicle, similar to pre-crash safety systems. However, even though stopping could avoid some types of failures, it might as well cause other failure cases such as rear-end crashes. Therefore, a good "safe" policy should consider the environmental situation and optimize towards a relatively safe action. 
This can actually be done by following a recently developed data-driven approach \cite{maeda2019reconnaissance} for constrained Markov Decision Process (CMDP). Given a baseline policy $\eta$, the greedy-safe policy $\pi^p_{\eta}$ is defined by
$\pi^p_{\eta}(U_i) = \argmin_{a} \mathscr{T}^\eta(U_i, a) $
where $\mathscr{T}^\eta(U_i, a)$ is the \emph{threat function} of $\eta$ given by
\begin{align}
    \mathscr{T}^\eta(U_i, a) = P( U^\eta \in \mathF \mid \text{initial action}= a )
\end{align}
with $U^\eta = T(\eta, (\pi^o, U_i, \mathcal F, \mathcal E), *)$ being the final state following $\eta$.
The threat function at $(U_i,a)$ is the probability for the ego vehicle to encounter a failure starting from the state $U_i$ with the initial action $a$ and then following the baseline policy $\eta$. Therefore, it can be learned by running the simulator with the baseline policy.
The threat function is similar to the Q-function in reinforcement learning, but instead of estimating the future reward, the threat function estimates the future chance of failure.
Similar to the greedy policy in reinforcement learning, the greedy-safe policy $\pi^p_{\eta}$ acts greedily according to the threat function to avoid failure.
Note that the greedy-safe policy $\pi^p_{\eta}$ is in a policy class different from the tested planner model $\pi^p_{\theta}$.

Using the greedy-safe policy, we define the generic avoidable failure as follows.
\begin{definition}[Generic avoidable failure]
Given a planner policy $\pi_{\theta}^p$ and a scenario $s = (\pi^o, U_{0}, \mathF, \mathE)$ such that $T(\pi_{\theta}^p, s, n) \in \mathF$,
$(\pi_{\theta}^p, s)$ is a \emph{generic avoidable failure} if and only if $T(\pi^p_{\eta}, s^{\prime}, *) \in \mathE \cap \mathF^{c}$ where $s^{\prime} = (\pi^o, T(\pi_{\theta}^p, s, n - \lceil\rho/\Delta t\rceil), \mathF, \mathE)$.
\end{definition}

Since this class of avoidable failures is determined by the greedy-safe policy $\pi^p_{\eta}$, they can provide planner-independent feedback for failure avoidance improvement. The choice of the baseline policy $\eta$ could affect the ability of the greedy-safe policy, but we show in our experiments that a random baseline policy can still detect significant generic avoidable failures.

\section{Planner Test Environment}
In this section, we describe the details of our planner testing settings: target traffic scenes and the tested planner.

\subsection{Traffic Scenes}\label{subsec:traffic-scenes}
\begin{figure}[t]
 \begin{minipage}[b]{0.48\linewidth}
  \centering
  \includegraphics[keepaspectratio, width=4.0cm]
  {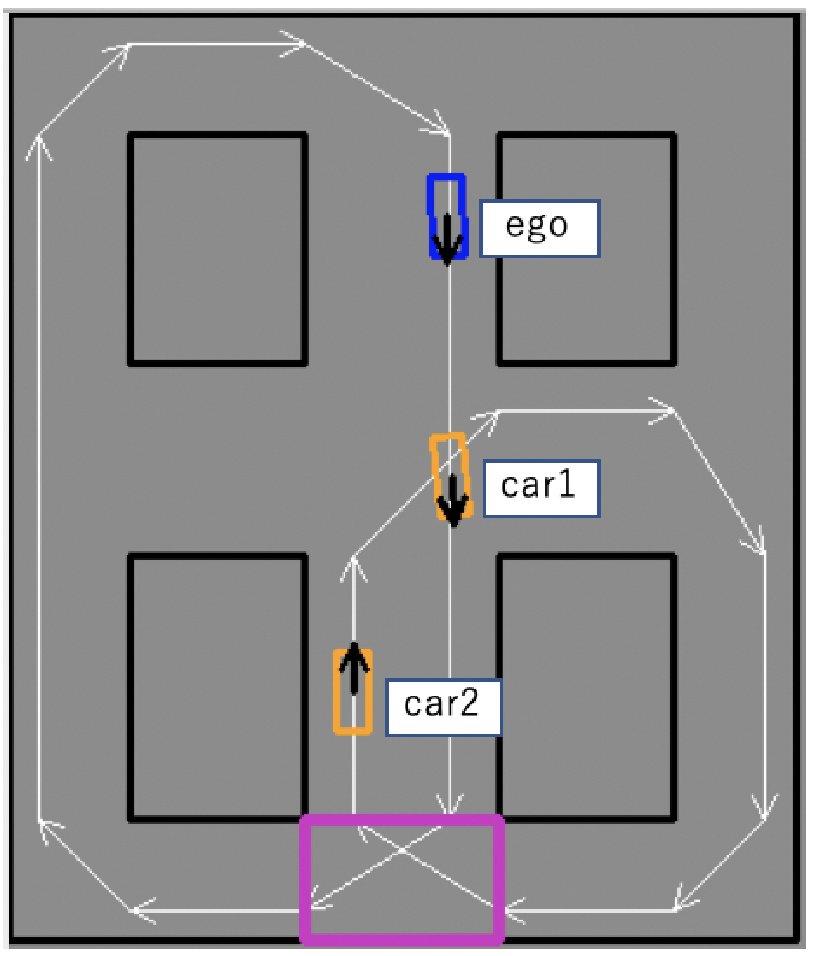}
  \subcaption{\textsf{Right-turn}}\label{traffic-scenes-right-turn}
 \end{minipage}
 \begin{minipage}[b]{0.48\linewidth}
  \centering
  \includegraphics[keepaspectratio, width=4.0cm]
  {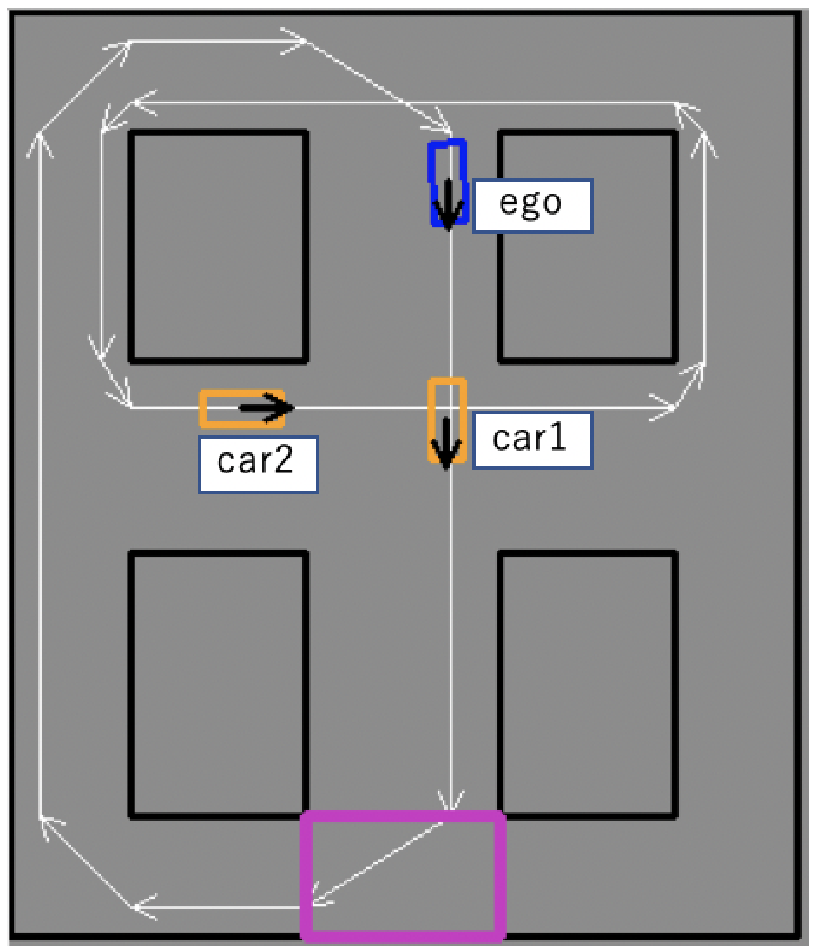}
  \subcaption{\textsf{Crossing}}\label{traffic-scenes-crossing}
 \end{minipage}
 \caption{The blue vehicle is the ego vehicle, and the orange vehicles are the other vehicles. 
 The label on each vehicle indicates the vehicle ID. White arrows indicate the course of each vehicle. Note that each vehicle is allowed to drive in places other than the white arrow on the course. The purple rectangle represents the goal of the ego vehicle, and it is considered a success case if the vehicle reaches the goal without collision and within designated time.}\label{traffic-scenes}
\end{figure}

We conduct planner testing assuming two types of traffic scenes in an intersection following Japanese traffic rules.
One (\textsf{Right-turn}) is a scene where the ego vehicle goes straight against the right-turning vehicle (Fig. \ref{traffic-scenes-right-turn}), and the other (\textsf{Crossing}) is a scene where the ego vehicle intersects another vehicle from the other lane (Fig. \ref{traffic-scenes-crossing}).
In both scenes, there is a leading vehicle in front of the ego vehicle.
There are no traffic lights in either scene.

\subsection{System under Test Model}\label{subsec:tested-planner}
Following previous studies on planner testing~\cite{adaptive-test1, adaptive-test2}, we use a planner with a variant of IDM~\cite{idm} as well\footnote{Note that \cite{adaptive-test1, adaptive-test2} do not describe the details of the planner. Thus, the algorithm may not be the same.}.
The specific control algorithm and the adjustable parameters for detecting avoidable failures are shown below.

\begin{figure}[t]
  \centering
  \includegraphics[keepaspectratio, width=6cm]
  {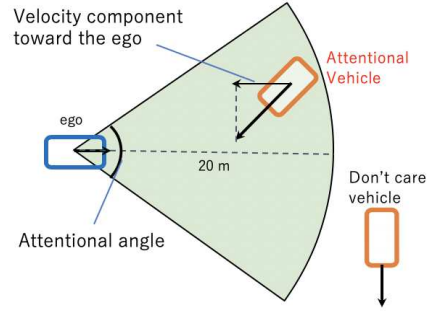}
  \caption{Longitudinal control of our tested planner. The ego car decelerates according to the other vehicle's velocity component toward the ego vehicle.}\label{longitudinal-control}
\end{figure}

Basically, the ego vehicle $g^e$ moves under IDM's free road behavior. If there is another vehicle around the ego vehicle, it decelerates according to that vehicle's speed.
In order to be able to run our planner in more involved traffic scenes, not limited to single-lane-following situations, such as intersections with multiple vehicles, we add a feature to select which vehicle to follow (we call it the \emph{attentional vehicle} here). The selection procedure is as follows.

First, the ego vehicle checks whether or not each of the surrounding other vehicles is included in the \emph{attentional area}, which is a $20~[m]$ and $\psi ~[\rad]$ fan-shaped area in front of the ego vehicle as shown in Fig.~\ref{longitudinal-control}.
The $\psi$ is called an \emph{attentional angle}, and dynamically changes in consideration of the possibility of collision with each vehicle.
For example, the attentional angle shrinks when the other vehicle is going-away.
The maximum and minimum attentional angles are determined by parameters $\phi_{max}$ and $\phi_{min}$.
The ego vehicle decelerates according to the closest attentional vehicle among the other vehicles determined to be included in the attentional area.
The desired deceleration is determined by the attentional vehicle's velocity component toward the ego vehicle and the parameter $d_{\text{IDM}}$  (see the original IDM~\cite{idm}).
With the aid of this selection rule, our IDM-based planner is able to run through intersections better, avoiding to awkwardly follow oncoming vehicles on the opposite lane while paying attention to crossing vehicles.

Especially in planner-dependent avoidable failure detection, a combination of the adjustable parameters of the planner related to the above longitudinal control is $\theta = (d_{\text{IDM}}, \phi_{max}, \phi_{min})$.

\section{Experiments}
In this section, we show the experimental results in detecting diverse failure cases that are relevant to the planner.

\subsection{Experimental Setting}
\subsubsection{Policy set for other vehicles}
To train the RL-based other vehicle policies, we generated training scenarios by applying random perturbations based on the vehicles' layout shown in Fig.~\ref{traffic-scenes}.
Using DPS (see \ref{subsec:generating-other-vehicle-policy}), we generated a combination of $50$ policies that have the highest behavioral diversity and a success rate of $90\%$ or more on the diversity evaluation scenarios generated with a different random number from the training.
We briefly denote the such diverse policy set as \textsf{Diverse}.

We also created a baseline policy set, called \textsf{LessDiverse}, with a low diversity score as follows:
i) create a set $\Pi_{cand}$ of many RL-based policies with success rate of $90\%$ or more,
ii) pick one policy $\pi^{o_k}$ from $\Pi_{cand}$,
iii) select $50$ policies from $\Pi_{cand}$ that are considered to be having similar behavior to $\pi^{o_k}$ based on their interpolicy-diversity.

For reference, the inter-policy diversity values of the two policy sets are shown in Table~\ref{tbl:policy-sets}. 
Note that a higher inter-policy diversity score is considered better in terms of diversity.

\begin{table}[t]
 \centering
 \caption{Prepared policy sets for other vehicles}
 \begin{tabular}{lllrr} \hline
      Traffic type & Policy type & Suc. rate   &  \scriptsize{$D_{IP}(\Pi_o)$} \\ \hline\hline
      \textsf{Right-turn} & \textsf{LessDiverse}  & 94.8 \% & 0.72 \\
      \textsf{Right-turn} & \textsf{Diverse}  & 94.0 \% & 5.11         \\ \hline
      \textsf{Crossing} & \textsf{LessDiverse} & 94.6 \% & 0.38         \\
      \textsf{Crossing} & \textsf{Diverse}  & 93.8 \% & 3.00
      \\ \hline
 \end{tabular}
 \label{tbl:policy-sets}
\end{table}

\subsubsection{Testing environment and scenarios}
The update time interval of a simulator $\Delta t = 0.1~[s]$.
As shown in \ref{subsec:traffic-scenes}, we conduct planner testing on two traffic scenes, which consist of two vehicles (\textsf{car1}, \textsf{car2}) other than the target ego vehicle. 
We prepared $300$ patterns by randomly introducing perturbations to the initial status of each vehicle.
These $300$ patterns are used to test every other vehicle driving policy.
Since we use $50$ different other driving policies, we performed $15,000$ simulation tests per each traffic scene.

In the tests, the simulation will end when the ego vehicle reaches the designated goal or when it fails to do so within $300$ steps.

A collision between the ego vehicle and other vehicles is also regarded as a termination condition, and the test result is counted as a failure.
When a failure due to a collision occurs, the proposed two avoidability verification methods are conducted to determine if the failure is avoidable or not.  

\subsubsection{Parameters of the tested planner}
We used an IDM-based planner as described in \ref{subsec:tested-planner}.
The adjustable parameters for testing are as follows: $d_{\text{IDM}} = 0.5~[m/s^2]$ which is the default setting of original IDM~\cite{idm}, $\phi_{max} = 3.142~[\rad]$ and $\phi_{min} = 0.524~ [\rad]$.
The following restrictions were given to ego vehicles:
maximum and minimum velocity $v_{max} = 2.0~[m/s]$, $v_{min} = -0.5~[m/s]$, and
maximum and minimum acceleration $a_{max} = 1.0~[m/s^2]$, $a_{min} = -1.0~[m/s^2]$.

\subsubsection{Parameter search to find planner-specific avoidable failure}

When a collision occurs, we evaluate if the collision is a planner-specific avoidable failure, defined in \ref{subsec:planner-dependent-AF}, by searching the three planner parameters shown in \ref{subsec:tested-planner}.
Since checking all possible combinations of these parameters is computationally infeasible, we make use of Optuna~\cite{optuna}, a parameter optimization framework, to run parameter search 100 times.
If there is a combination of parameters which succeeds in avoiding the evaluated collision, the collision is counted as a planner-specific avoidable failure.
We set the parameter space of the tested planner as $\Theta_p = \{(d_{\text{IDM}}, \phi_{max}, \phi_{min})~|~d_{\text{IDM}} \in [0.1, 10.0], \phi_{max} \in [\frac{\pi}{2}, 2\pi], \phi_{min} \in [\frac{\pi}{12}, \frac{\pi}{2}]\}$.
We also set the minimum reaction time $\rho$ of ego vehicle to $2.0~[s]$ based on \cite{driver-reaction-time}.

\subsubsection{Greedy-safe policy for generic avoidable failure}
When a collision is not avoidable by the planner-specific modification, we evaluate if the collision is a generic avoidable failure, by running the greedy-safe policy defined in \ref{subsec:planner-independent-AF}.
We use the policy which selects completely random actions as the baseline policy $\eta$ to construct the threat function. For computational efficiency, we approximate the threat function by neural networks with the threat function upper bound in \cite{maeda2019reconnaissance}. 
Since the actions take continuous values, in our experiments, the greedy-safe policy $\pi^p_\eta$ randomly samples $30$ actions and pick the one with the lowest threat value at each time.
We also set the minimum reaction time of the ego vehicle $\rho = 2.0~[s]$.

\subsection{Counting Avoidable Failures}

Table~\ref{tbl:count-collision} shows the number of failures detected in each of combinations of traffic types and policy types.
Here we classify the failures into three types; planner-specific avoidable failures, generic avoidable failures found only by the planner-independent modification, and unavoidable failures. 
We can see that the planner-specific modification can identify a decent amount of avoidable failures, and these types of failures can be directly used in adjusting the parameter of the planner policy. For those failures unavoidable by the planner-specific modification, the proposed planner-independent method provides additional failure feedback to planner development. These types of generic avoidable failures could help indicate certain weaknesses of the target policy class, and they provide information about the general kinetic capabilities of the ego vehicle. The rest of unavoidable failures are likely irrelevant to the planner design, as discussed in previous sections.


The ability of policy diversity in finding failures is obvious from Table~\ref{tbl:count-collision}; the total number of failures found by the \textsf{Diverse} is more than that by the \textsf{LessDiverse}, about $1.5$ times more in the \textsf{Right-turn} scene and about $10$ times more in the \textsf{Crossing} scene.
Moreover, many failures found by the \textsf{LessDiverse} are unavoidable, about one quarter, $28/100$, in the \textsf{Right-turn} scene and about half, $11/23$, in the \textsf{Crossing} scene. While for the diverse policy set, only about one-tenth, $15/153$, of the failures in the \textsf{Right-turn} scene and about one quarter, $80/235$, in the \textsf{Crossing} scene are unavoidable. This difference shows a major advantage of using diverse policies in planner testing. 

There is one interesting observation for avoidable failures that the proportions of planner-specific avoidable failures of the two sets of policies are actually similar, all-around half of the total number of failures. The proportion is $52/100$ for \textsf{LessDiverse} and $75/153$ for \textsf{Diverse} in the \textsf{Right-turn} scene, and is $12/23$ for \textsf{LessDiverse} and $102/235$ for \textsf{Diverse} in the \textsf{Crossing} scene. However, the number of generic failures found by the \textsf{LessDiverse} is far less, even none was detected in the \textsf{Crossing} scene.
Therefore, another major benefit of the diverse policy sets is the capability to produce generic avoidable failures regardless of the target policy model.


\begin{table}[t]
 \centering
 \caption{The number of each failure types:\\
 \textbf{Total}:~total \#failures in 15,000 simulation tests, \textbf{A-P}:~~\#planner-specific avoidable failures, \textbf{A-G}:~\#generic avoidable failures detected by the planner-independent modification only,
 \textbf{U}:~\#unavoidable failures.
 }
 \begin{tabular}{ll|r|rrr}\hline
      & & &  \multicolumn{3}{|c}{Failure types}\\
      Traffic type & Policy type & \textbf{Total} & \textbf{A-P} & \textbf{A-G} & \textbf{U} \\\hline\hline
      \textsf{Right-turn} & \textsf{LessDiverse} & 100 & 52 & 20 & 28 \\
      \textsf{Right-turn} & \textsf{Diverse}     & 153 & 75 & 63 & 15 \\ \hline
      \textsf{Crossing} & \textsf{LessDiverse}   &  23 & 12 &  0 & 11 \\
      \textsf{Crossing} & \textsf{Diverse}       & 235 & 102 & 53 & 80 \\ \hline
 \end{tabular}
 \label{tbl:count-collision}
\end{table}

\subsection{Diversity of Detected Failures}
In this subsection, we will examine the planner testing results from three different aspects: colliding vehicles, collision angles, and collision positions. 

First, colliding vehicles. Table~\ref{tbl:classification-collided-car} classifies the avoidable collisions based on which vehicle (\textsf{car1}, \textsf{car2}) the ego-vehicle collided with. 
These avoidable collisions are found through avoidability analysis, either by the planner-specific method or the generic detection.
The results show a larger number of collisions with \textsf{car2}, representing an oncoming vehicle in the \textsf{Right-turn} scene, and a vehicle coming from the side in the \textsf{Crossing} scene, comparing to \textsf{car1}, which precedes ego-vehicle in the same lane on both scenes and much less likely to collide with.
We can see that the tests using diverse policies were able to induce collisions with \textsf{car1}, which could be avoidable, while testing with less diverse policies was not able to detect these types of collisions.

\begin{figure*}
 \begin{minipage}[b]{0.2\linewidth}
  \centering
  \includegraphics[keepaspectratio, width=3.5cm]
  {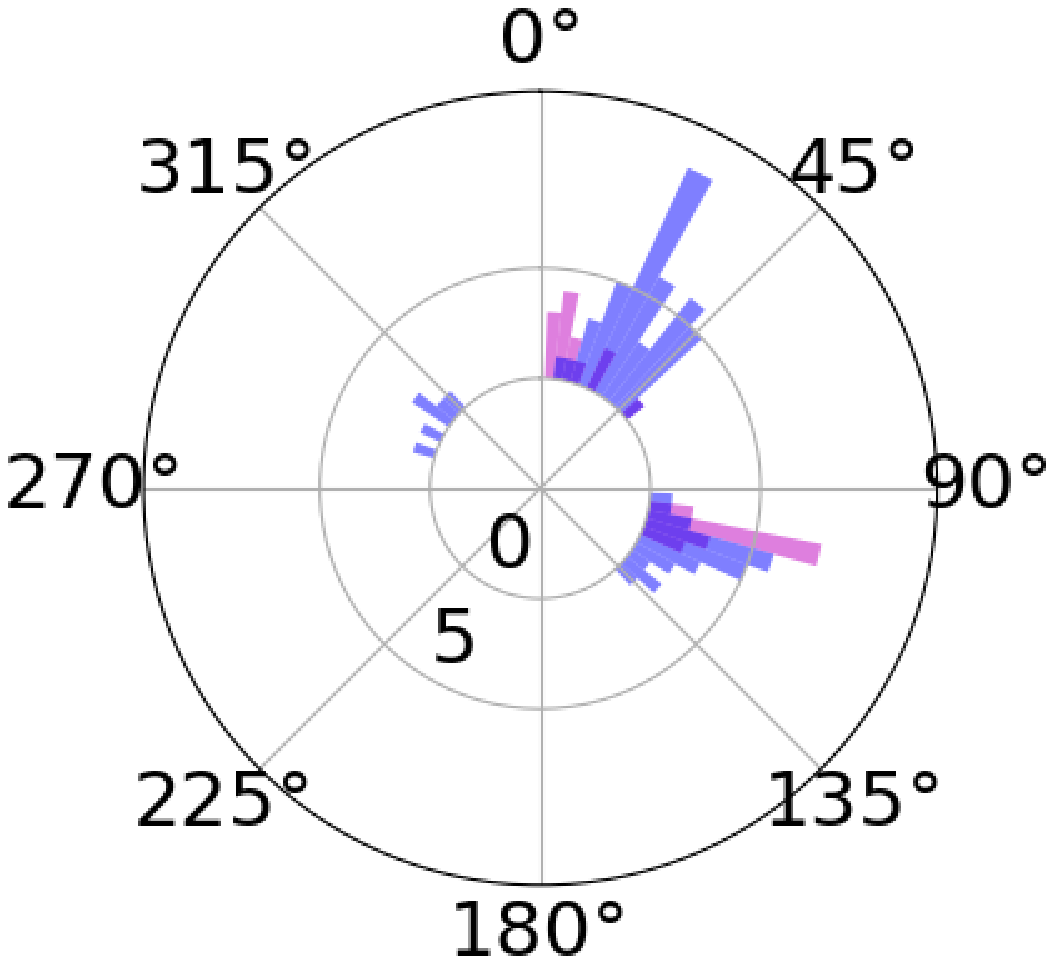}
  \subcaption{\textsf{LessDiverse}, \textsf{Right-turn}}
 \end{minipage}
 \begin{minipage}[b]{0.2\linewidth}
  \centering
  \includegraphics[keepaspectratio, width=3.5cm]
  {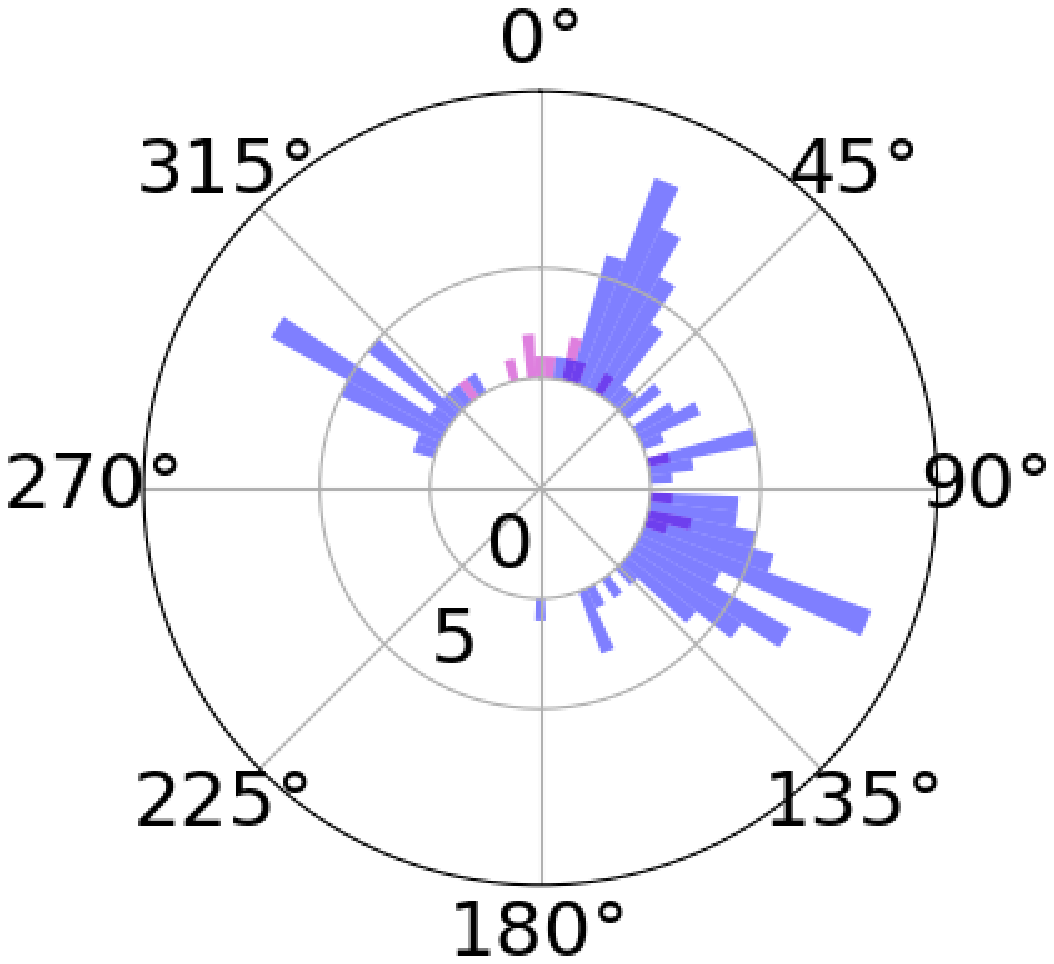}
  \subcaption{\textsf{Diverse}, \textsf{Right-turn}}
 \end{minipage}
 \begin{minipage}[b]{0.2\linewidth}
  \centering
  \includegraphics[keepaspectratio, width=3.5cm]
  {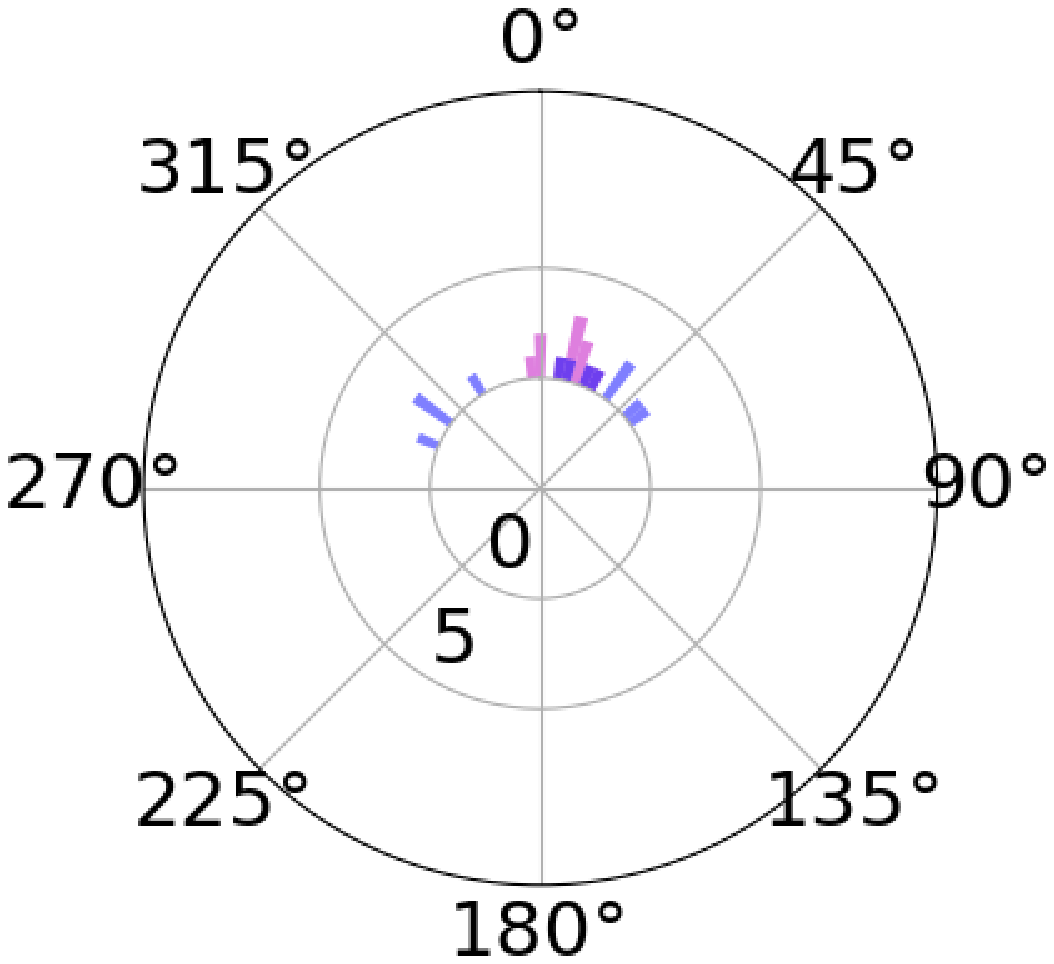}
  \subcaption{\textsf{LessDiverse}, \textsf{Crossing}}
 \end{minipage}
 \begin{minipage}[b]{0.2\linewidth}
  \centering
  \includegraphics[keepaspectratio, width=3.5cm]
  {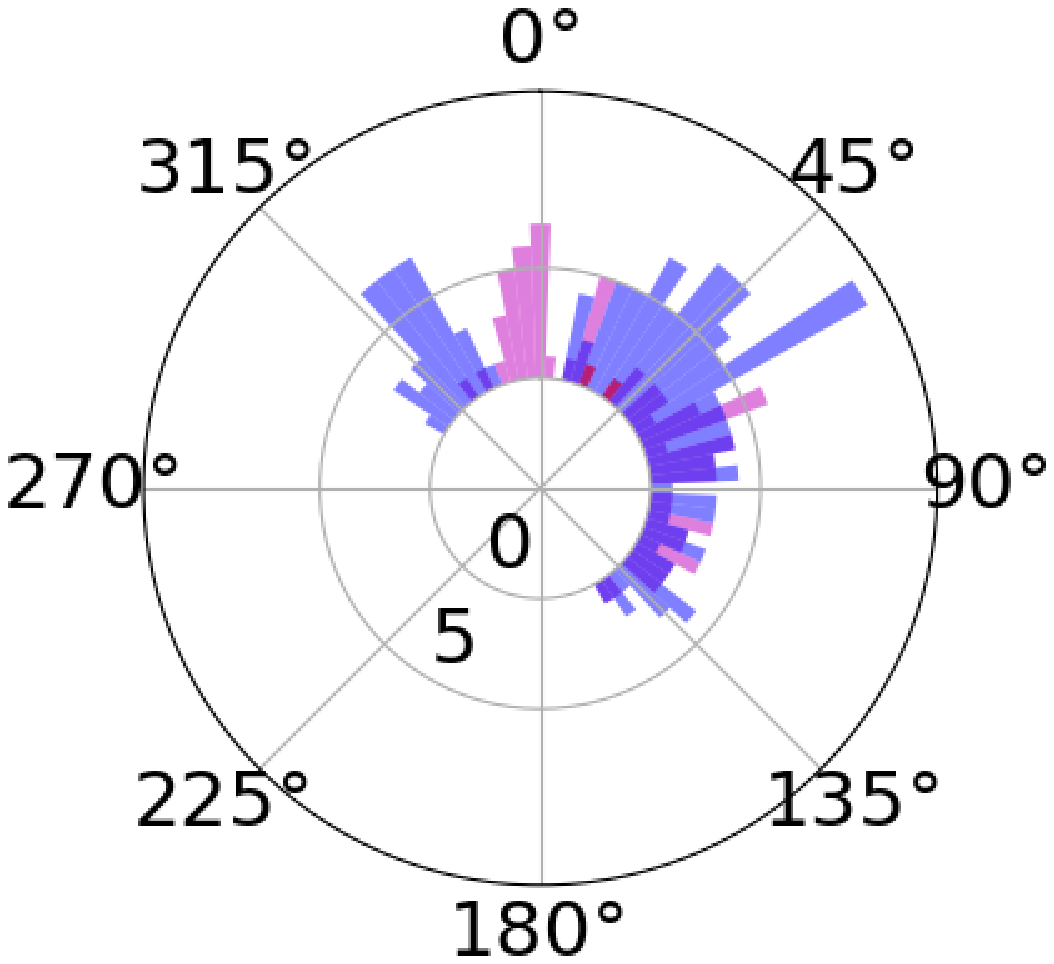}
  \subcaption{\textsf{Diverse}, \textsf{Crossing}}
 \end{minipage}
  \begin{minipage}[b]{0.18\linewidth}
  \centering
  \includegraphics[keepaspectratio, width=3.0cm]
  {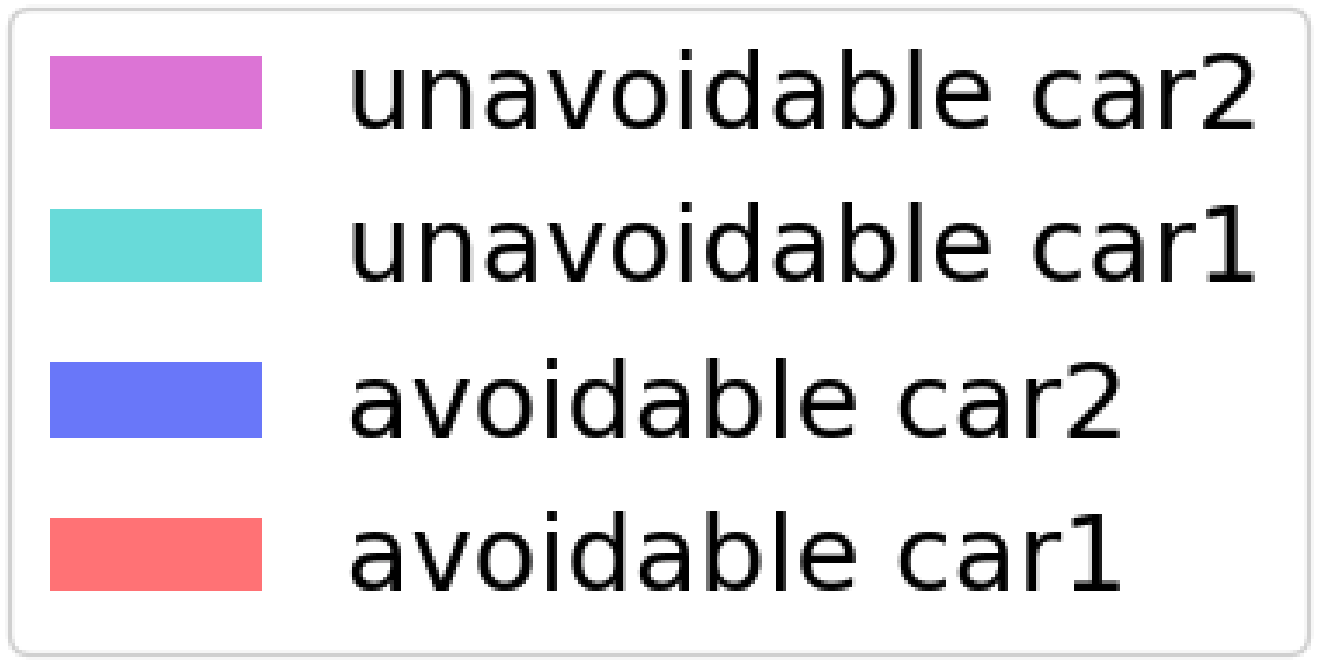}
 \end{minipage}
  \caption{The distribution of collision angle. The number of  the bins is 80.}\label{fig:collision-angle-diversity}
\end{figure*}

Secondly, Fig. \ref{fig:collision-angle-diversity} displays the distribution of collision angles, where the ego-vehicle angle is considered as 0$^{\circ}$, reference at the time of collisions. We can clearly see that the set of diverse policies results in collisions with a much wider distribution of collision angles.

\begin{figure*}
 \begin{minipage}[b]{0.2\linewidth}
  \centering
  \includegraphics[keepaspectratio, width=3.5cm]
  {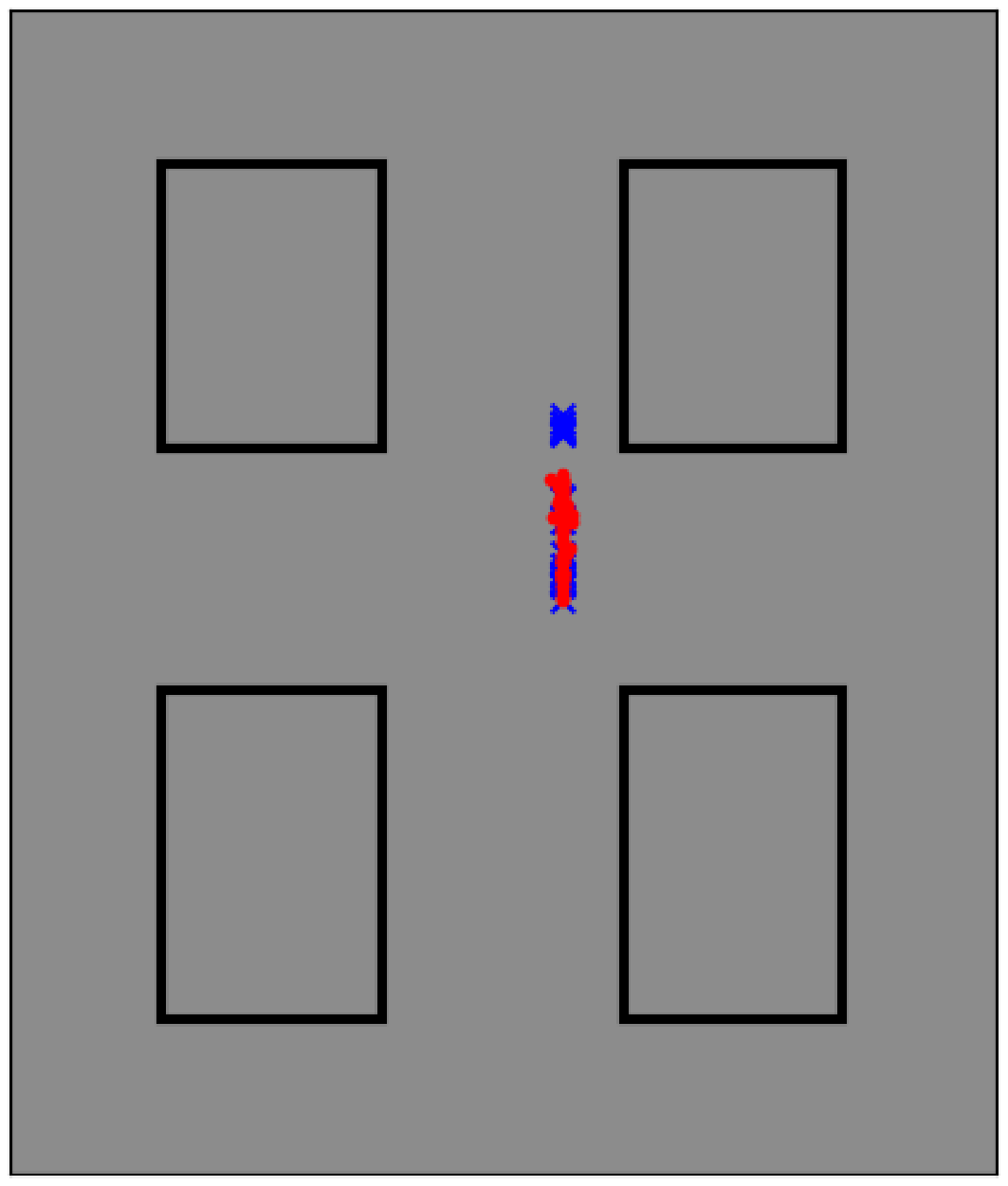}
  \subcaption{\textsf{LessDiverse}, \textsf{Right-turn}}
 \end{minipage}
 \begin{minipage}[b]{0.2\linewidth}
  \centering
  \includegraphics[keepaspectratio, width=3.5cm]
  {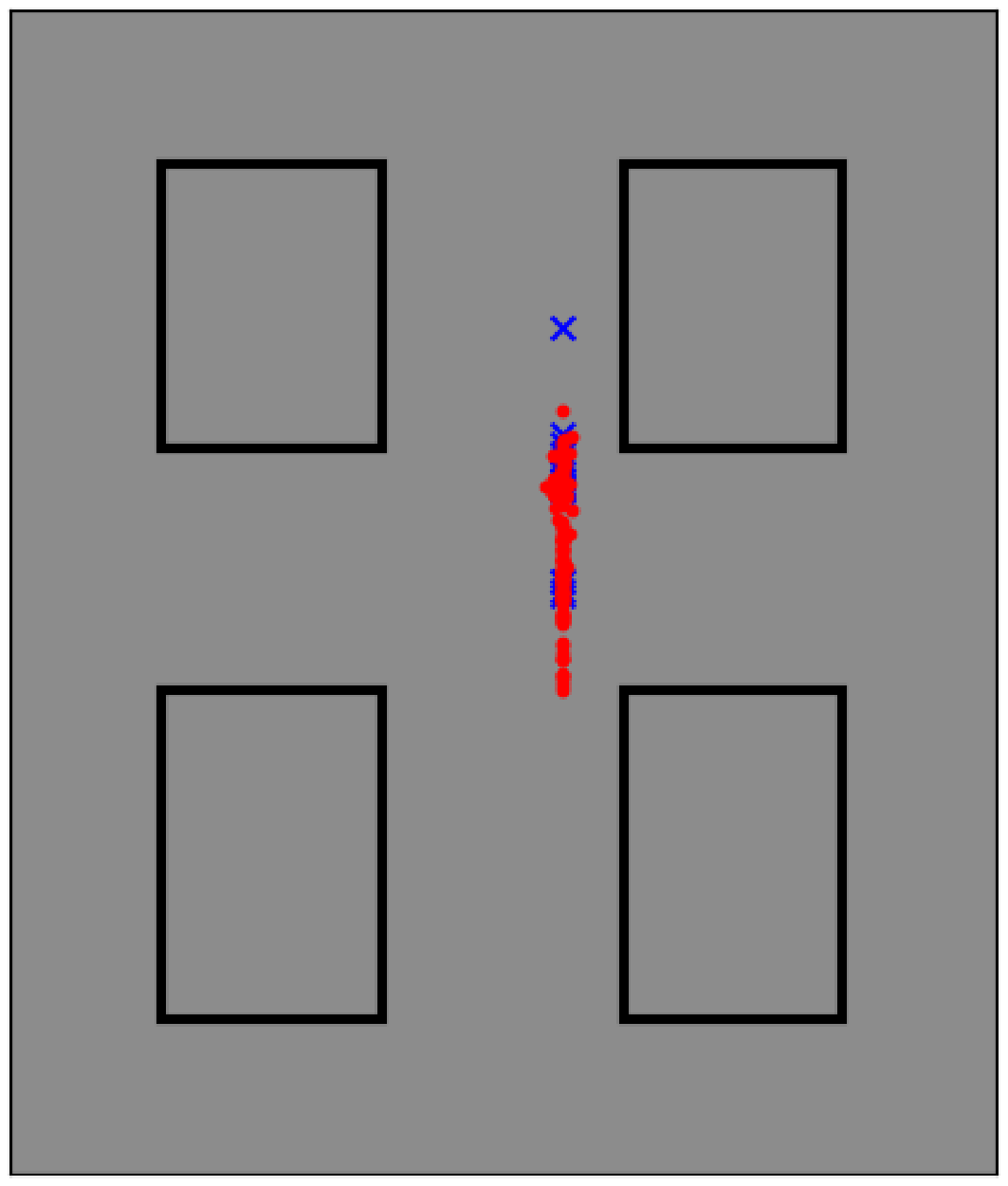}
  \subcaption{\textsf{Diverse}, \textsf{Right-turn}}
 \end{minipage}
 \begin{minipage}[b]{0.2\linewidth}
  \centering
  \includegraphics[keepaspectratio, width=3.5cm]
  {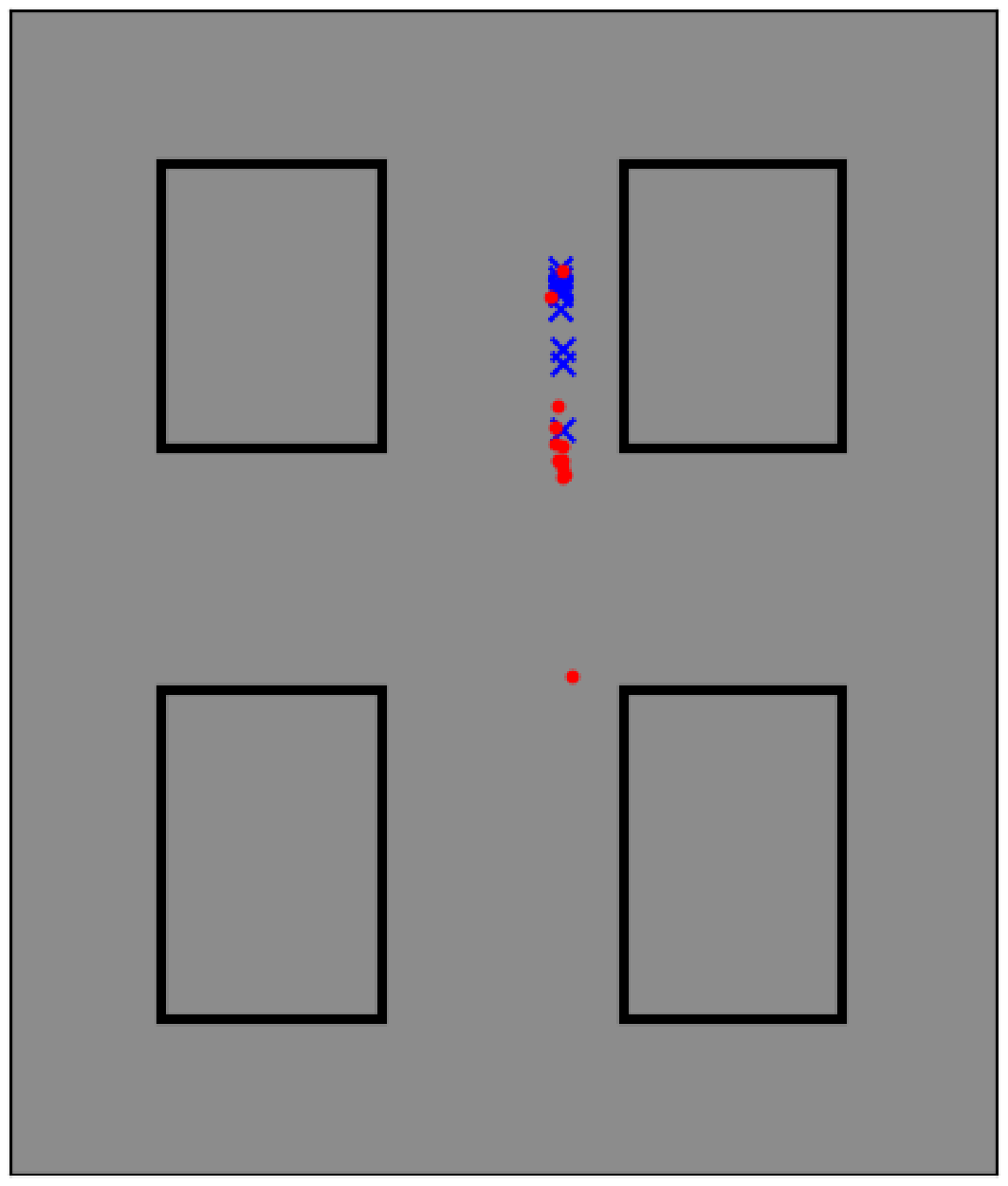}
  \subcaption{\textsf{LessDiverse}, \textsf{Crossing}}
 \end{minipage}
 \begin{minipage}[b]{0.2\linewidth}
  \centering
  \includegraphics[keepaspectratio, width=3.5cm]
  {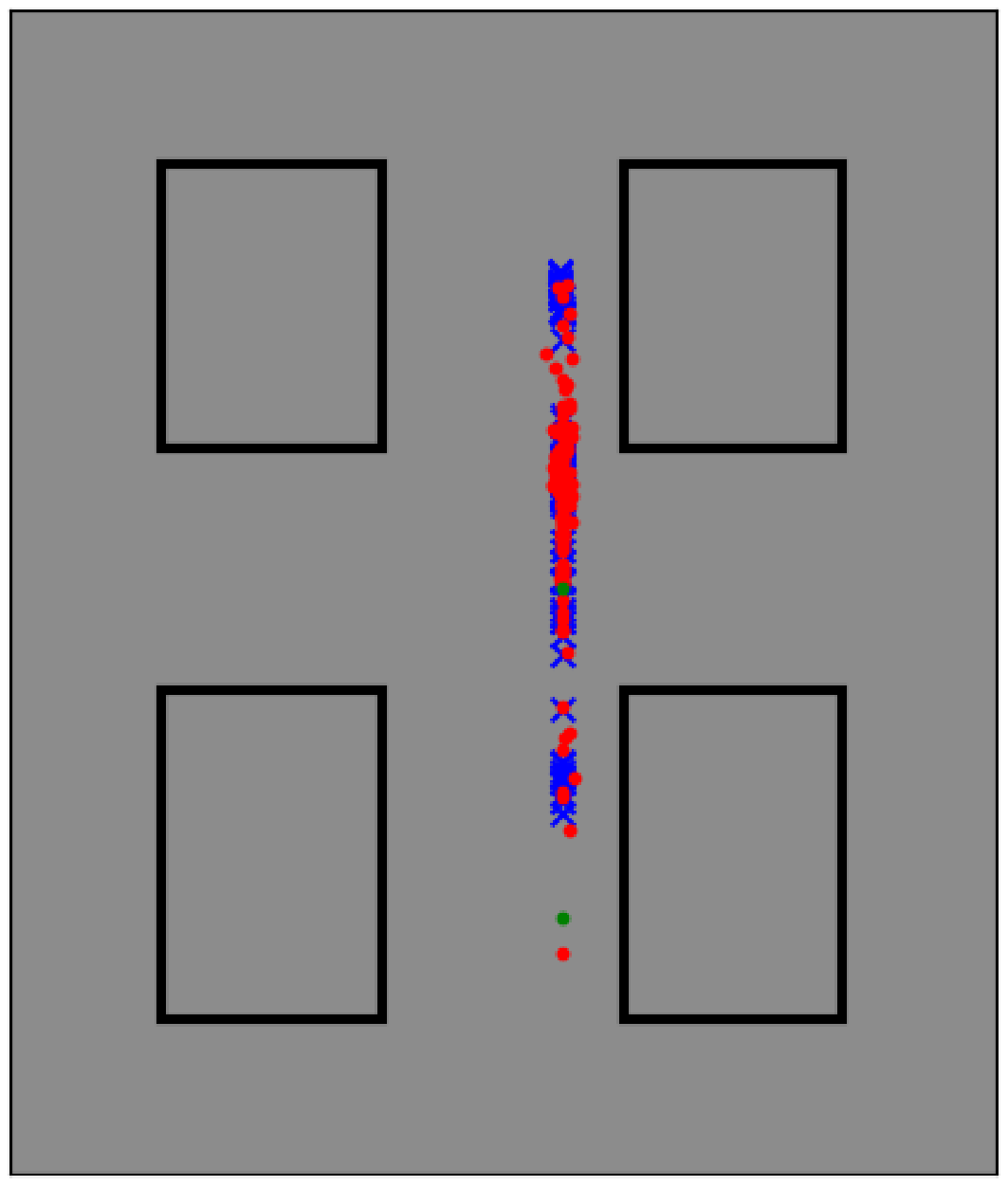}
  \subcaption{\textsf{Diverse}, \textsf{Crossing}}
 \end{minipage}
  \begin{minipage}[b]{0.18\linewidth}
  \centering
  \includegraphics[keepaspectratio, width=3.0cm]
  {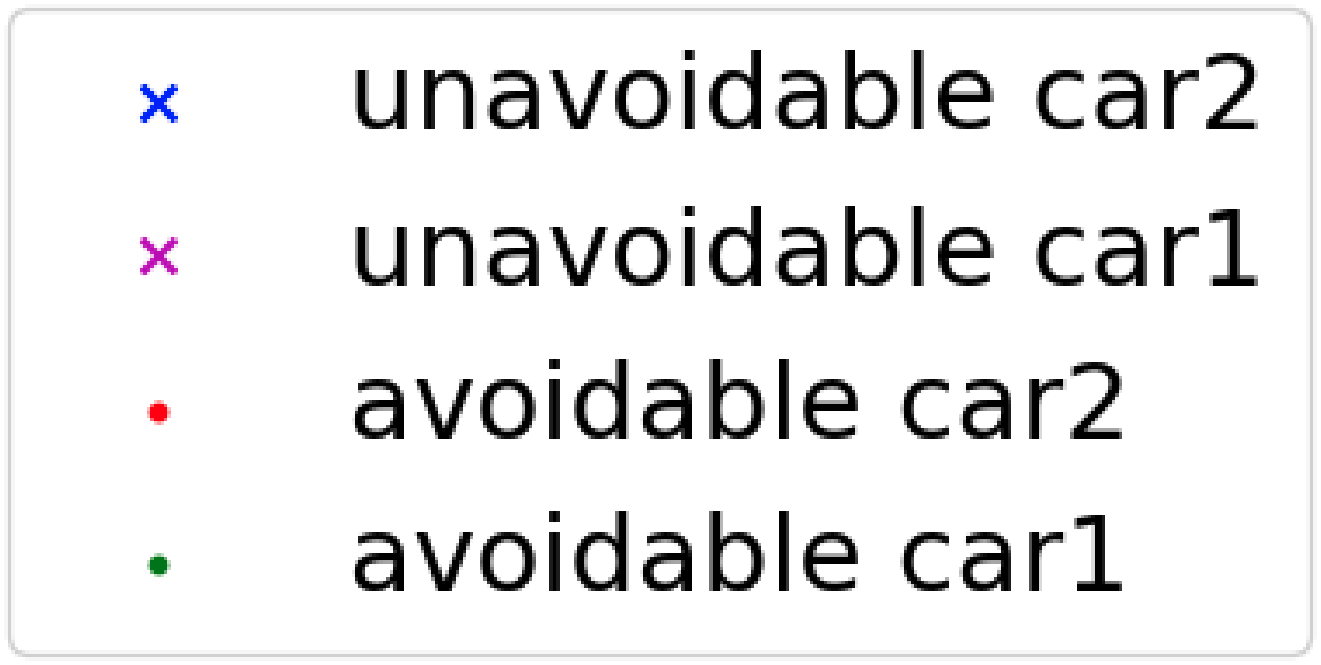}
 \end{minipage}
  \caption{The distribution of collision location.}\label{fig:collision-location-diversity}
\end{figure*}

Thirdly,  Fig. \ref{fig:collision-location-diversity} shows the distribution of collision locations in the traffic simulation map. Since both tested driving scenes target the straight line itinerary, as the IDM-based planner also does not control steering, the resulted distributions of collision locations spread out linearly. Still, the set of diverse policies causes collisions with much wider distributions of collision positions.

The resulted distribution of collision locations also shows a group of collisions occurring in the early region of the test course (top middle location in the map), which were all unavoidable collisions. Examining these test scenarios more closely showed that the collisions were due to \textsf{car2} strangely leaving the designated course and aggressively heading towards the ego-vehicle, similar to what adversarial test scenarios generally aim for. While such cases are still rare, they do shade light on one difficulty in utilizing RL-based agents as other vehicles in traffic simulation. Fortunately, our collision avoidability verification check enables us to classify these test scenarios as unavoidable collisions, which could be considered of lower priority for investigation from the perspectives of planner development.

\begin{figure}[th]
 \centering
 \begin{minipage}[b]{0.48\linewidth}
  \includegraphics[keepaspectratio, width=4.5cm]
  {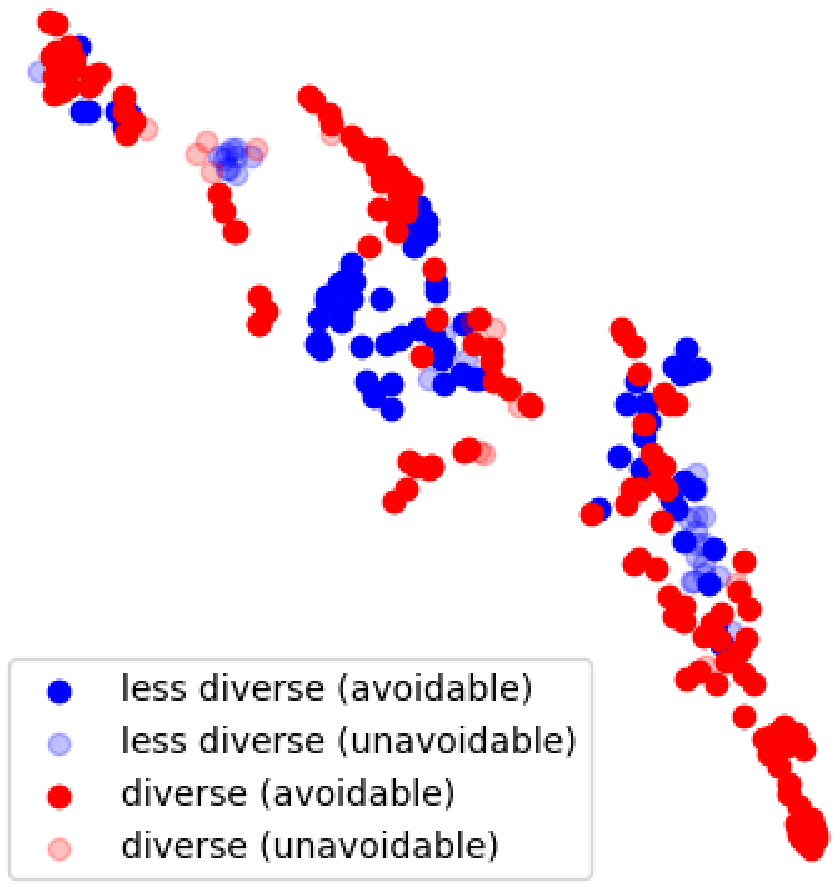}
  \subcaption{\textsf{Right-turn}}\label{fig:tsne-right-turn}
 \end{minipage}
 \begin{minipage}[b]{0.48\linewidth}
  \includegraphics[keepaspectratio, width=4.5cm]
  {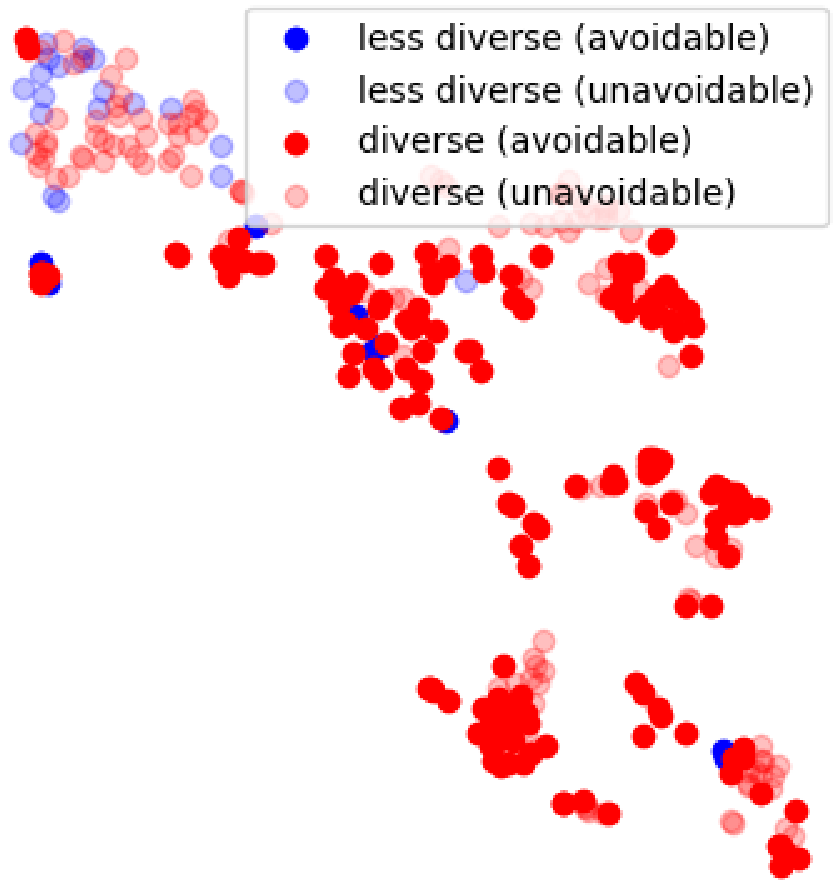}
  \subcaption{\textsf{Crossing}}\label{fig:tsne-crossing}
 \end{minipage}
 \caption{Visualization of the distribution of collision state by embedding with t-SNE.}\label{fig:tsne-visualization}
\end{figure}

Finally, we performed an integrated analysis combining the position, speed, angle, acceleration, and steering angle of the two vehicles involved in the collision.
Since this is a high-dimensional data analysis, we embedded the collision states into a two-dimensional space using t-SNE~\cite{t-sne} as shown in Fig.~\ref{fig:tsne-visualization}.
From the figure, we can see that the avoidable failures (dark red dots) resulting from the diverse policies have a wider distribution than the avoidable collisions (dark blue dots) resulting from the less diverse policies in the \textsf{Right-turn} scene.
For \textsf{Crossing}, the number of avoidable collisions for the less diverse is small, while we can see that the avoidable collisions resulting from the diverse policies display several clusters.

\begin{table}[t]
 \centering
 \caption{The number of avoidable collision for each vehicle (collision to \textsf{car1}/\textsf{car2}) of failure cases.}
 \begin{tabular}{ll|rr} \hline
                     & &  \multicolumn{2}{|c}{\#Collision} \\
      Traffic scene & Policy type & \textsf{car1} & \textsf{car2} \\\hline\hline
      \textsf{Right-turn} & \textsf{LessDiverse} &    0 &   72 \\
      \textsf{Right-turn} & \textsf{Diverse}     &    0 &  138 \\ \hline
      \textsf{Crossing} & \textsf{LessDiverse}   &    0 &   12 \\ 
      \textsf{Crossing} & \textsf{Diverse}       &    2 &  153 \\ \hline
 \end{tabular}
 \label{tbl:classification-collided-car}
\end{table}

\subsection{Planner Weakness Found Through Testing}
\begin{figure}
 \begin{minipage}[b]{0.24\linewidth}
  \centering
  \includegraphics[keepaspectratio, width=2.0cm]
  {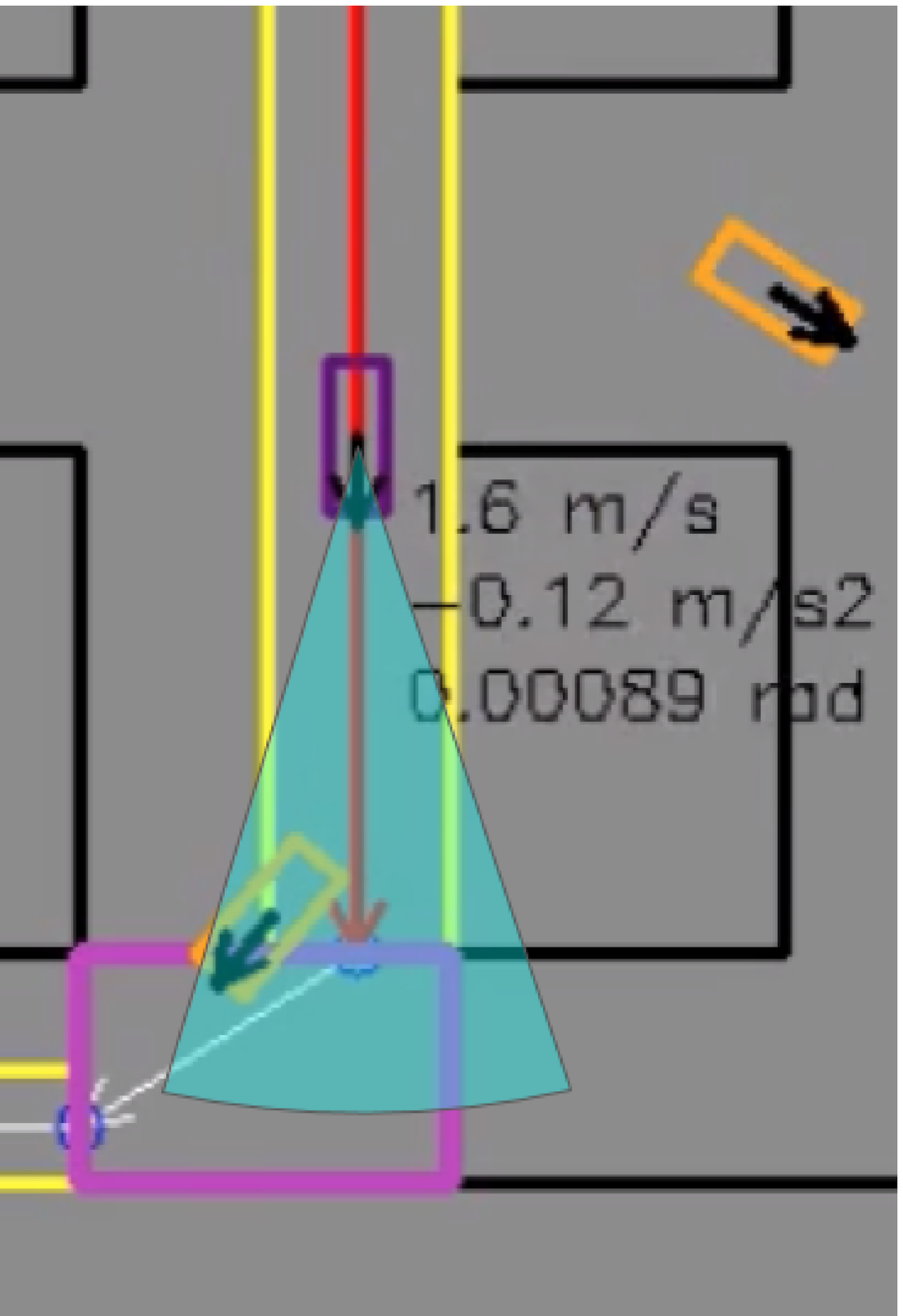}
  \subcaption{}\label{fig:planner-weakness-a}
 \end{minipage}
 \begin{minipage}[b]{0.24\linewidth}
  \centering
  \includegraphics[keepaspectratio, width=2.0cm]
  {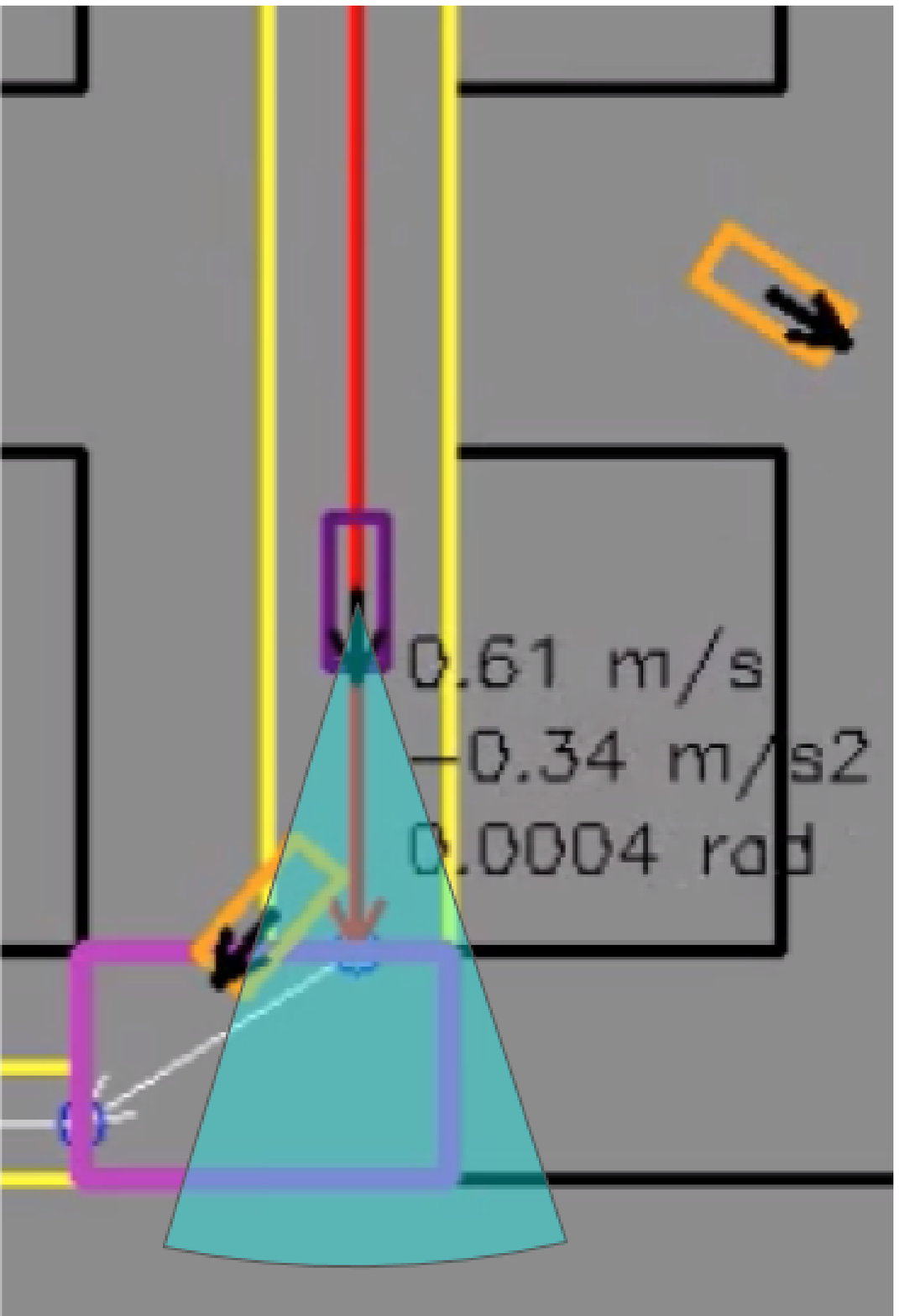}
  \subcaption{}\label{fig:planner-weakness-b}
 \end{minipage}
 \begin{minipage}[b]{0.24\linewidth}
  \centering
  \includegraphics[keepaspectratio, width=2.0cm]
  {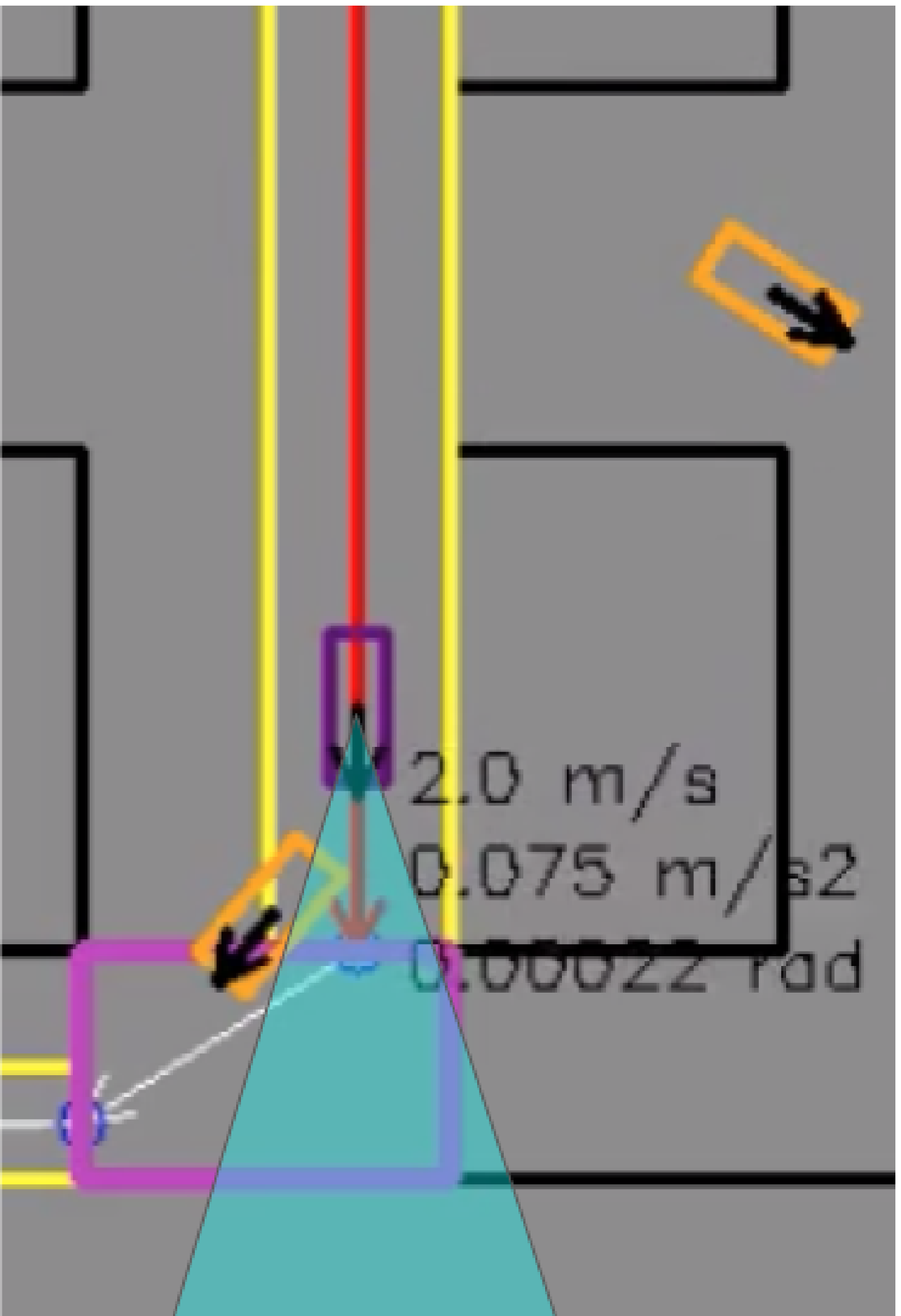}
  \subcaption{}\label{fig:planner-weakness-c}
 \end{minipage}
 \begin{minipage}[b]{0.24\linewidth}
  \centering
  \includegraphics[keepaspectratio, width=2.0cm]
  {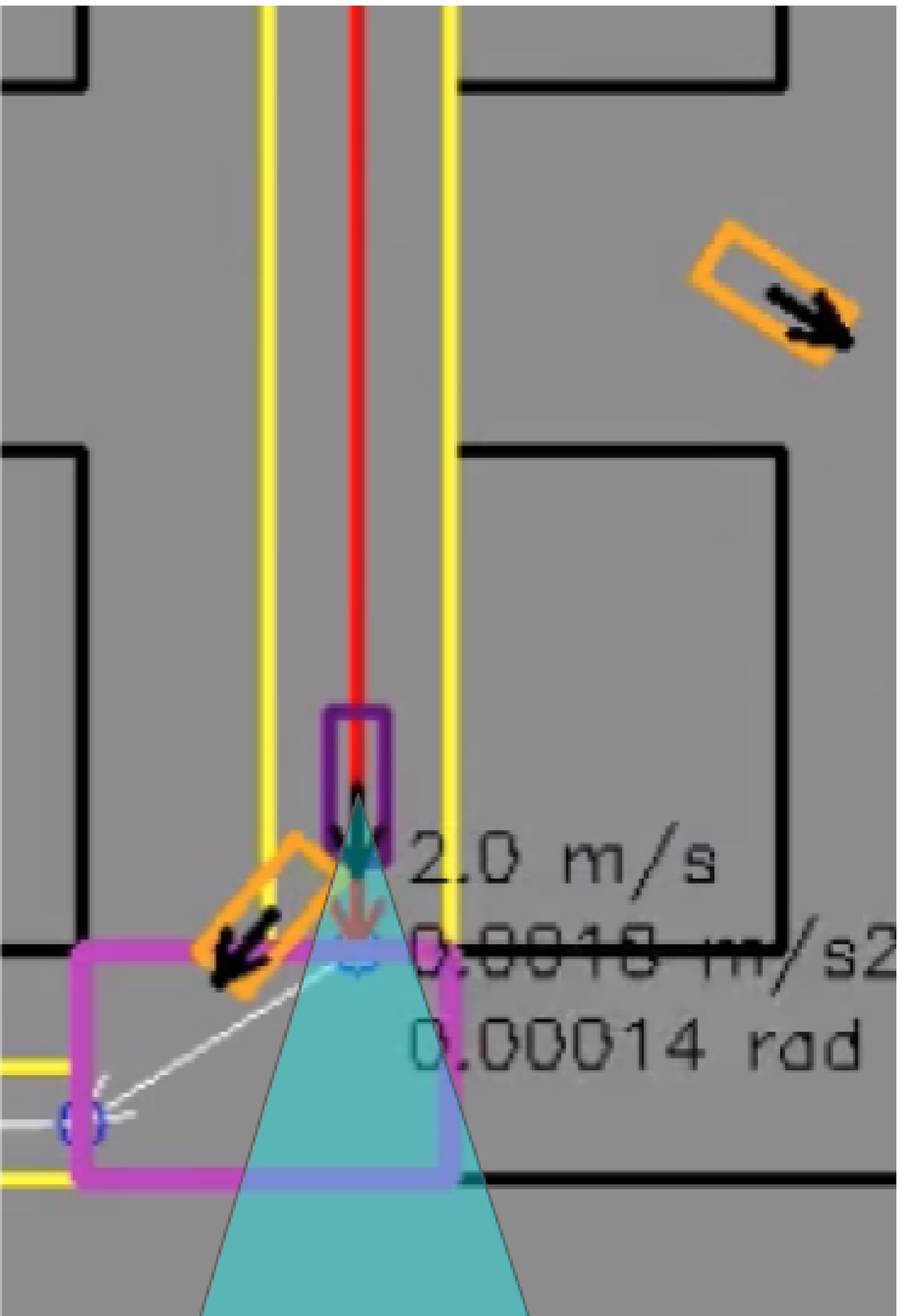}
  \subcaption{}\label{fig:planner-weakness-d}
 \end{minipage}
  \caption{An example of failure cases of the IDM-based planner.}\label{fig:planner-weakness}
\end{figure}

We examined the avoidable failures and found one of the failure cases indeed reflects a weakness of the IDM-based planner. Fig. \ref{fig:planner-weakness} illustrates the situation where the leading vehicle (\textsf{car1}) suddenly stops at the corner on the bottom of the map when taking a turn (\ref{fig:planner-weakness-a}). The ego vehicle first applies a brake to avoid a collision (\ref{fig:planner-weakness-b}), but at a certain timing, it starts accelerating again (\ref{fig:planner-weakness-c}) and eventually crashes into the leading vehicle (\ref{fig:planner-weakness-d}). This is due to the planner's limited attentional angle $\psi = \phi_{min}$ according to the attentional angle computation. This is a planner-specific avoidable failure, and increasing the attentional angle should have avoided this particular collision. However it could also impact the IDM behavior in other situations, and the trade-off is to be assessed in the development cycle based on these tests.

\subsection{Discussion on Near-miss Cases}

\begin{figure*}[t!]
 \begin{minipage}[b]{0.24\linewidth}
  \centering
  \includegraphics[keepaspectratio, width=4.2cm]
  {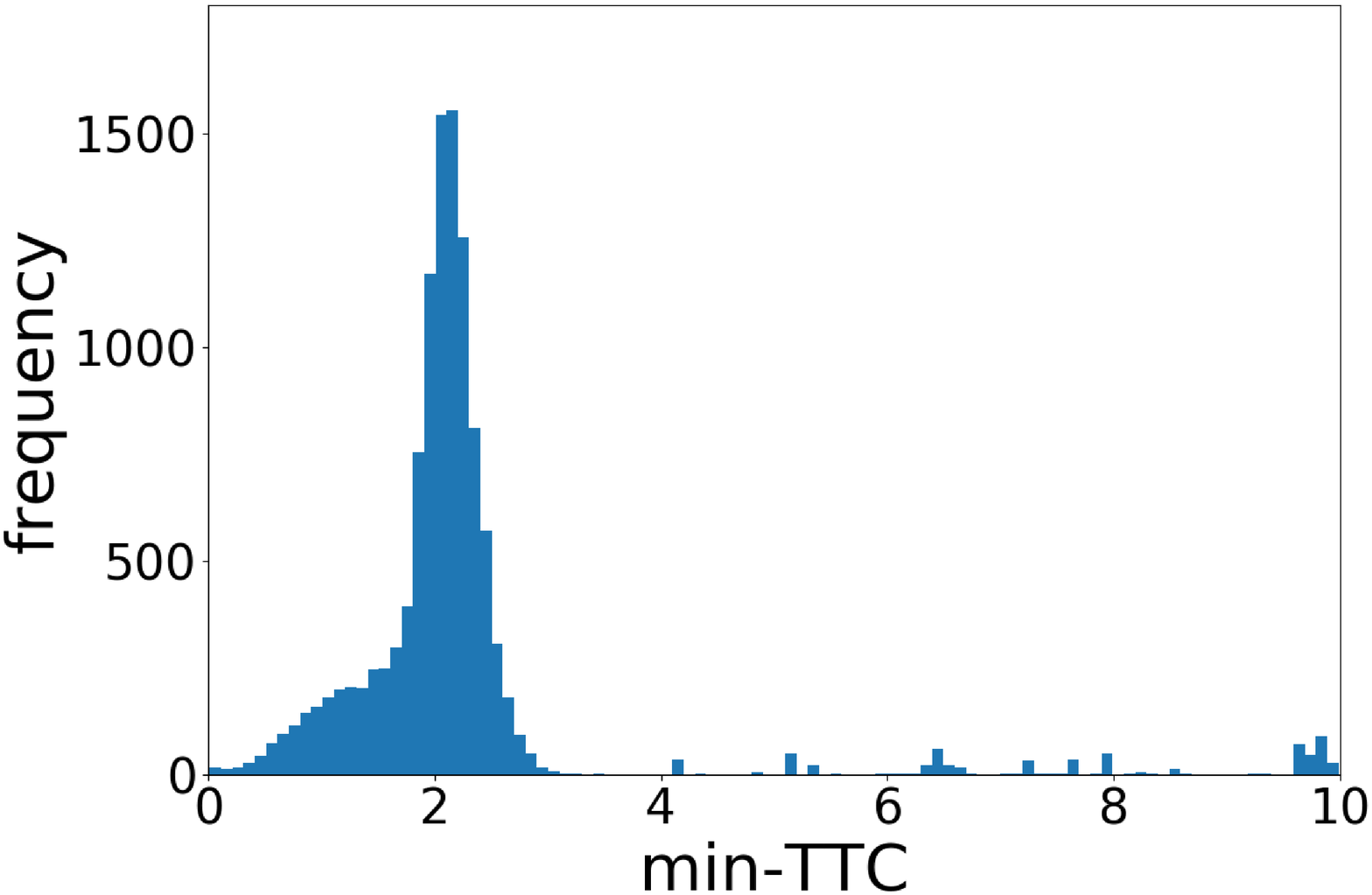}
  \subcaption{\textsf{LessDiverse}, \textsf{Right-turn}}
 \end{minipage}
 \begin{minipage}[b]{0.24\linewidth}
  \centering
  \includegraphics[keepaspectratio, width=4.2cm]
  {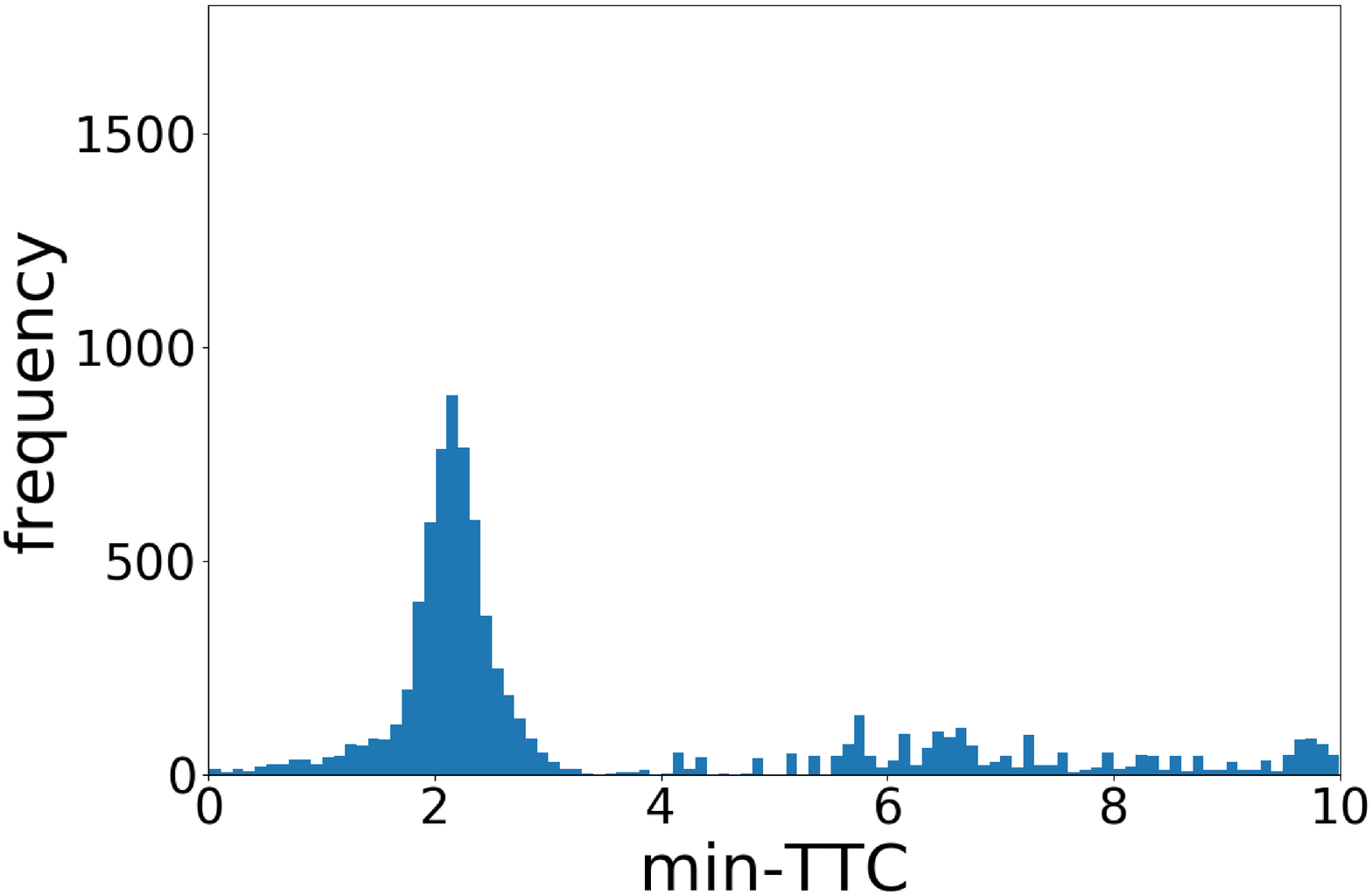}
  \subcaption{\textsf{Diverse}, \textsf{Right-turn}}
 \end{minipage}
 \begin{minipage}[b]{0.24\linewidth}
  \centering
  \includegraphics[keepaspectratio, width=4.2cm]
  {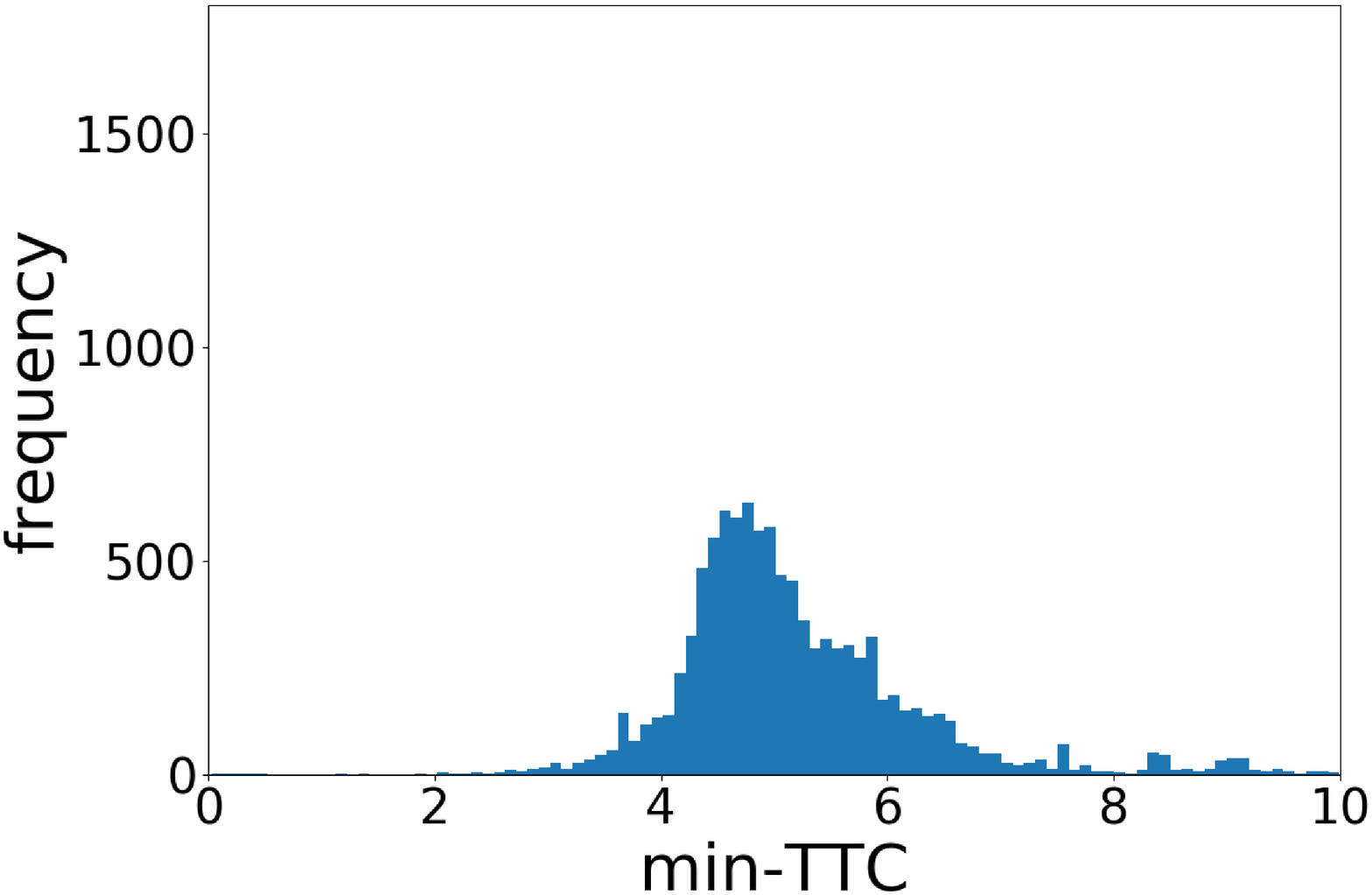}
  \subcaption{\textsf{LessDiverse}, \textsf{Crossing}}
 \end{minipage}
 \begin{minipage}[b]{0.24\linewidth}
  \centering
  \includegraphics[keepaspectratio, width=4.2cm]
  {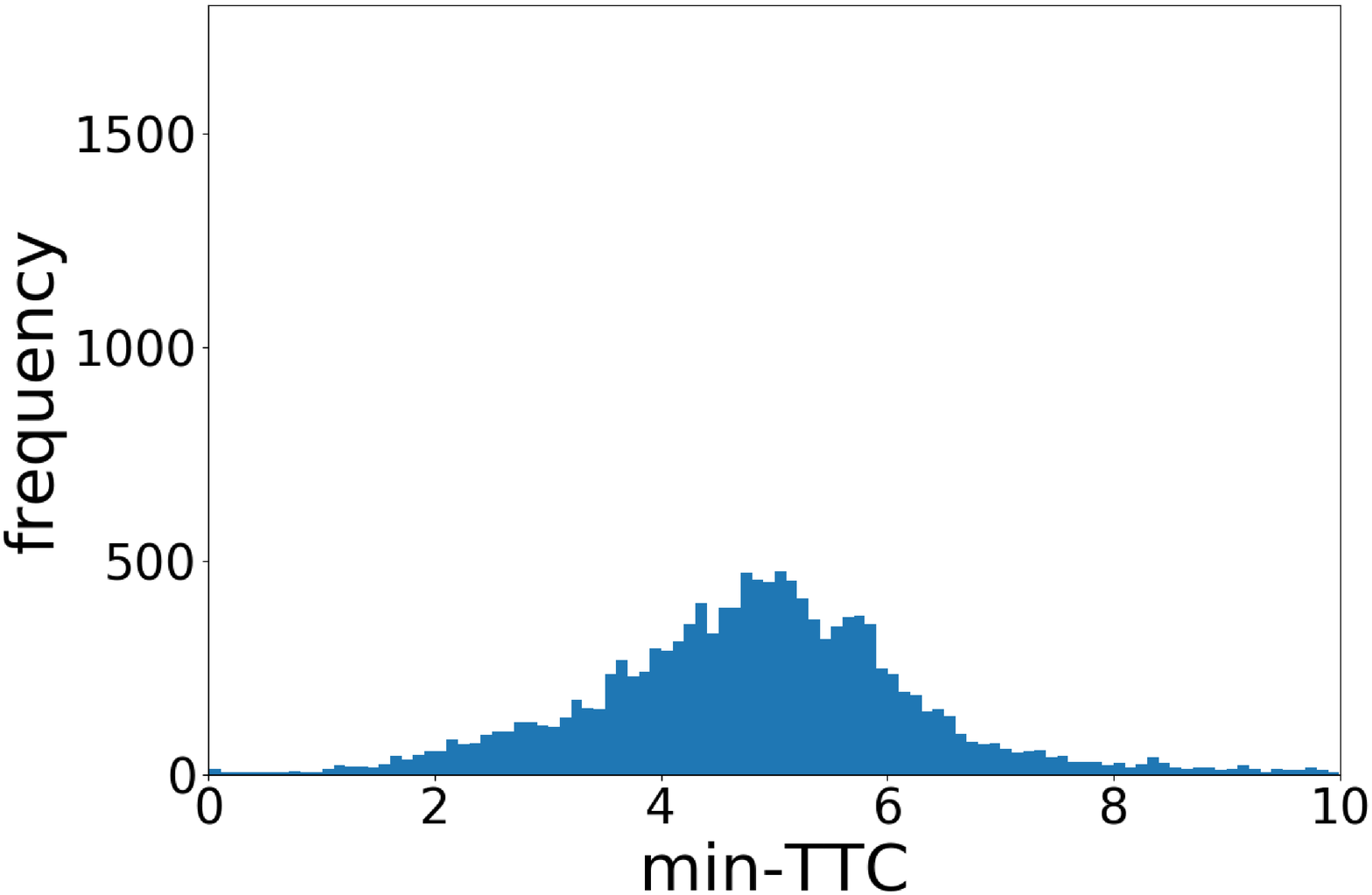}
  \subcaption{\textsf{Diverse}, \textsf{Crossing}}
 \end{minipage}
   \caption{The distribution of min-TTC. The number of the bins is 100.}\label{fig:ttc-diversity}
\end{figure*}

In addition to collisions, near-miss situations are also of great interest in planner testing, as they can sometimes indicate errors or malfunctions in the system.
We implemented a two-dimensional Time To Collision (TTC)~\cite{computing_ttc} as a preliminary experiment to count the number of minimum TTCs (min-TTCs), where min-TTC is defined as the smallest TTC between the ego vehicle and all other vehicles through a simulation test.

Fig.~\ref{fig:ttc-diversity} shows distributions of the obtained min-TTCs for each policy set on both traffic scenes.
Comparing the distributions of diverse policies and less diverse policies, we can see that the distributions are spread in both scenes.
In the \textsf{Crossing} scene, more near-miss situations of $\text{min-TTC} \leq 2.0~[s]$ were detected with diverse policies testing.
However, in the case of \textsf{Right-turn}, the result was the opposite, most probably because the behaviors of the less diverse policies are biased towards that.
In other words, various situations covering min-TTC can be created by using diverse policies.
Usually, if a planner developer wants to test such a lower min-TTC case, they have to design other vehicles manually to produce such a targeted situation.
For example, in the case of using RL-based policies, they would need to give an agent a reward when it produces a lower min-TTC.

In addition, we checked by replaying the near-miss cases in the \textsf{Right-turn} with less diverse policies.
We found many cases that the ego vehicle was yielding to the right-turn vehicle which turned right without slowing down.
This near-miss case, when the planner policy has already taken a safe behavior, is not necessarily important for planner testing.
Therefore, we believe that it is necessary to define near-miss cases that are relevant to planner testing and to narrow down the analysis of near-miss cases to relevant ones.
It is not appropriate to apply the proposed avoidable failure detections in this paper to near-miss detection because the planner policy does not always need to avoid such near-misses if the ego vehicle is already taking safe behavior.
In the future, it is a challenge to detect relevant near-miss cases, including the improvement of two-dimensional TTC.

\section{Conclusion}
In this work, we experimentally demonstrated a simple but scalable approach to detecting various relevant planner failure cases, using behaviorally diverse policies for traffic simulation vehicles.
In the test results analysis, we used our two proposed failure avoidability assessment mechanisms to counterfactually identify avoidable failures, being important feedback from planner testing.
We also examined the resulted failure test cases from multiple angles, and were able to demonstrate that the proposed approach helped detect a wider range of avoidable planner failure situations. 
While the tests were conducted on an IDM-based planner that has limited flexibility, we do expect that this approach shall result in much more diverse classes of detected failures when applied on more practical planning systems.

Furthermore, the diversity capacity of the simulation environment itself dictates the scale of diverse behaviors possible within it. Such capacity can be expanded by adding more maps, traffic lights, pedestrians, different vehicle types, etc. to cover as many scenes as the automated vehicles would need to navigate the real world.
Also, if we can automatically generate various traffic maps and rules, we will be able to build our testing environment more efficiently.
We do expect that our proposed framework shall demonstrate increasing benefit as such capacity increases, moving automated vehicles a step forward towards a robust real-world deployment.

\section*{Acknowledgment}
This work was supported by Toyota Research Institute Advanced Development, Inc.

\bibliographystyle{IEEEtran}
\bibliography{bibtex}

\end{document}